**Designing an efficient and equitable humanitarian supply chain dynamically via reinforcement learning**

Author: Weijia Jin

# 1. Introduction

Unexpected events like natural disasters, extreme weather events induced by climate change, geopolitical events, and pandemics devastate living environments and lives. The 2025 California wildfires resulted in at least twenty-seven fatalities, forced over two hundred thousand residents to evacuate, and destroyed or damaged more than twelve thousand four hundred structures. Although humanitarian supply chains (HSCs) should sustain the essential supply to mitigate adverse effects caused by large-scale devastation (Timperio et al., 2022), the efficiency of the HSCs has been undermined by the issues of cutting the budget to local fire departments, thus particularly affecting resource allocation and emergency management that causes widespread dissatisfaction according to public discourse surrounding the 2025 California wildfires (Erokhin, 2025).

Moreover, except for prioritizing the paramount concern of efficiency in HSCs under emergency circumstances, the equitable allocation of resources is the revelation brought by the recent years of the COVID-19 pandemic. For instance, as of September 2021, less than 4% of Africa's population had received at least one dose of a COVID-19 vaccine, while the vaccination rate in Europe and North America exceeded 70%. By early 2022, the U.S. had discarded over 82 million doses of COVID-19 vaccines due to declining demand and vaccine expiration. These phenomena as oversupply in certain areas and insufficient aid for marginalized or remote communities had prevailed during pandemic times.

HSCs can be classified into short-term and long-term types. While short-term HSCs address urgent needs during acute disasters, long-term HSCs support sustainable recovery by tackling issues like infrastructure rebuilding, disease management, and food insecurity (Yılmaz et al., 2025). From the perspective of the international situation in recent years, most require long-term humanitarian supply chains, making the relationship between efficiency and cost growingly important while accounting for equity as a necessary factor.

Nevertheless, there is relatively limited research that integrates efficiency, equity and cost simultaneously. Common practice is to embed the goal of improving efficiency or meeting demand at various demand points within the goal of cost minimization, or split these goals into the multi-objective problem, then rely on the preferences to select the solutions from the Pareto front.

On the other side, many studies formulate the problem as two-stage stochastic programming models with static settings under different scenarios, such as disruptions and demand uncertainty in the HSCs field (Yılmaz et al., 2025). Whereas, in reality, the demand points of each demand point of the impacted population are dynamically changing. Moreover, there is seldom practice using machine learning or reinforcement learning in HSCs since such problems sometimes include binary and discrete variables at the same time, traditionally, is suitable for linear programming or simple heuristic algorithms. However, given the growing frequency of crises, it is critical to incorporate dynamic analysis with a high-performance algorithm, reinforcement learning (RL), to provide more realistic meanings and implications for HSCs.

Motivated by these, this study seeks to answer the following research questions (RQs):

RQs-1: How can efficiency, equity, and cost be integrated into dynamic HSCs optimization?

RQs-2: How does the RL adapt to the dynamic optimization?

RQs-3: How does the model perform under different environmental settings?

RQs-4: How is RL superior to the traditionally adopted algorithms in HSCs?

This study makes novel contributions from the following perspectives:

An attempt to apply theory to algorithms: A dynamic RL model is applied under the theoretical framework of cost-efficiency with equity concern. Additionally, the attempt to monetize efficiency that can be seen in the methodology is a rare approach in similar studies.

The construction of optimization problem objectives: The study combines the two objectives together using RL model, not just as common linear programming methods or heuristic algorithms usually use a single objective, like minimizing the cost or maximizing the profit, or selecting solutions from the Pareto front when adopting multi-objective approaches as mentioned, for optimization.

New developments in the use of RL: Drawing from previous experience, RL cannot learn effectively within complex logics and formulations, but in this study, a hybrid of binary and discrete proximal policy optimization (PPO), which is one kind of RL algorithm, is used for logistics optimization. Previously, mainly academic papers used RL in supply chain management that particularly concentrates on inventory management, but this time the study is a comprehensive analysis encompassing the reward, the level of meeting demands, cost, efficiency, and inventory with the ultimate goal to upgrade distribution efficiently and equitably.

Horizontal comparisons: The study provides the heuristic single-objective optimisation algorithm, particle swarm optimisation (PSO), and multi-objective optimisation algorithms, non-dominated sorting genetic algorithm II (NSGA-II), for comparison purposes.

In addition, training AI agent to make decisions whether virtually or physically can provide a fast response when a crisis happens or for extreme situations that no human can quickly make wiser decisions. Meanwhile, it is more intelligent than some logistics software since the AI agent processes its own observations, to a certain extent, its own "thinking".

The remainder of this study is organized as follows: Section 2 gives the literature review on HSCs design. Section 3 presents the problem description and the dynamic optimization model. Section 4 displays the experiments' setting, key results, and corresponding explanations. Finally, Section 5 provides the concluding remarks.

## 2. Literature review

HSCs are involved with various nodes to assist the recovery and supply necessities for impacted people in challenging periods. They have developed to provide rapid disaster support, focusing on planning, procurement, storage, and distribution to efficiently deliver resources (Haavisto and Kovács, 2015; Noham and Tzur, 2018). That requires the designing of HSCs considering the multi-objectives and even conflicting goals for different stages. Meanwhile, optimization methods such as linear programming (LP), mixed-integer programming (MIIP), stochastic programming (SP), heuristic methods as genetic algorithms (GA) and PSO have been widely used (Modarresi and Maleki, 2023; Wang et al., 2023; Yılmaz et al., 2025), but often struggle with dynamic and uncertain environments.

## 2.1 The focus and structure of HSCs design

Many researchers designed the various optimization purposes of the different parts of the HSC models by separating stages, phases, or multi-objective approaches. Kaur and Singh (2022) formulated two mathematical models for proactive and reactive situations to ensure an uninterrupted and resilient supply in HSC for three phases. Yılmaz et al. (2025) proposed a two-stage stochastic programming model that integrates GA and random forest (RF) to address demand and capacity uncertainties in humanitarian supply chains (HSCs), with findings indicating that incorporating machine learning into optimization methods enhances both resilience and long-term sustainability across different risk levels. Torabi et al. (2018) proposed a two-stage mixed fuzzy-stochastic programming model for integrated relief pre-positioning and procurement planning in humanitarian supply chains. Ershadi and Shemirani (2022) proposed a multi-objective optimization model for humanitarian relief logistics, focusing on minimizing unsatisfied injured people and optimizing transportation activities. Khalili-Fard et al. (2024) developed a bi-objective humanitarian supply chain model that optimizes response time and total cost while integrating governmental and non-governmental collaboration, pre-positioning, quantity flexibility contracts, and multi-sourcing policies to enhance logistical resilience and reliability.

Only a few studies have examined how to simultaneously optimize HSC responsibilities efficiently and equitably. Modarresi and Maleki (2023), leverage mixed-integer linear programming (MILP) models to design an equitable humanitarian relief supply chain by optimizing cost reduction and enhanced responsiveness. Wang et al. (2023) propose a multi-objective hyper-heuristic (MOHH) algorithm to address the location-allocation problem for emergency supplies, balancing timeliness and fairness in resource distribution. In the fairness-oriented research, Khorsi et al. (2020) and Cheng et al. (2021), both prioritize equitable distribution by incorporating social vulnerability metrics and dynamic allocation strategies. Khorsi et al. (2020) proposed a multi-objective MILP model for equitable relief item distribution, aiming to minimize total relief time, reduce unmet demands, and enhance fairness in allocation by utilizing the epsilon-constraint method to solve the optimization model. Similarly, Cheng et al. (2021) introduced a robust goal programming approach to ensure fair food distribution in humanitarian relief supply chains (HRSCs).

Some papers utilize reinforcement learning to improve the decision-making problem in HSCs. Yu et al. (2021) developed a reinforcement learning-based Q-learning algorithm to optimize resource allocation in humanitarian logistics. Their approach, formulated as a mixed-integer nonlinear programming model, balances efficiency, effectiveness, and equity while outperforming traditional heuristic and dynamic programming methods. Van Steenbergen et al. (2023) proposed a multi-trip, split-delivery vehicle routing model for last-mile humanitarian relief distribution using Unmanned Aerial Vehicles (UAVs) to mitigate travel time uncertainty, and demonstrated that dynamic reinforcement learning (RL) as value function approximation (VFA) and policy function approximation (PFA), improves performance and robustness compared to static solutions. Despite advancements, challenges remain in managing trade-offs between efficiency and fairness.

## 2.3 The cost-efficiency theory
Integrating efficiency and equity in humanitarian supply chains (HSCs) is crucial yet challenging, as traditional or general efficiency aims to ensure timely responses, while equity emphasizes equitable resource allocation to underserved populations that may require great evaluation time.

These two objectives seem contradictory to each other, and relevant organisations such as governments or charities also need to manage the cost of resource allocations within budgetary constraints subject to HSC management. Given the dilemma faced by HSC management, Cost-Benefit Analysis (CBA) is an approach that can combine the two objectives to achieve a feasible solution within the premise of improving the satisfaction level as much as possible while costs are controlled.

Early versions of CBA originated from the work of Jules Dupuit (1844/1952), who explored the idea of evaluating public infrastructure projects by comparing their costs and benefits. The approach was then formalized by Alfred Marsha in his subsequent works (Wiener, 2013). CBA can estimate and assess the value of a project or decision's benefits and costs to determine in both monetary and intangible forms and is widely utilized in various industries. Cost-efficient analysis evolved from this. The paper by Puett (2019) discusses cost-efficiency in humanitarian interventions by analyzing program costs of Action Against Hunger relative to output measures such as the number of beneficiaries served or the quantity of aid delivered. Maghsoudi et al. (2021) provide a systematic review of cash and voucher assistance (CVA) programs in humanitarian supply chains, emphasizing their potential to enhance cost-efficiency by reducing logistical burdens and improving aid responsiveness. There is no research conducted on this RL algorithm under the cost-efficiency framework in this field.

## 2.4 Background of PPO

PPO is an RL algorithm for training an intelligent agent to make decisions. Specifically, it is a policy gradient method, often used for deep learning when the policy network is very large. The predecessor to PPO, Trust Region Policy Optimization (TRPO), was published in 2015 by Schulman et al . It addressed the instability issue of another algorithm, the Deep Q-Network (DQN). PPO was published in 2017 by Schulman et al . It was essentially an approximation of TRPO that does not require computing the Hessian. The KL divergence constraint was approximated by simply clipping the policy gradient. Based on the trust region learning proposed by Schulman et al. (2015), proximal policy optimization (PPO) outperforms the previous DRL algorithms by guaranteeing monotonic policy improvement in every training iteration. The successor, such as heterogeneous-agent proximal policy optimization (HAPPO), a multi-agent deep reinforcement learning (MADRL) with decentralized approaches, exhibits outstanding performance in managing inventory level (Liu et al., 2024). However, the approach of that algorithm, such as decentralized decision-making or autonomous decision-making by multiple agents, is not the most ideal method to seek solutions in the context of humanitarian supply chains, which require centralized control and strong coordination under dynamical uncertainty.

## 3. Problem description and PPO in HSC optimization model

### 3.1 PPO Algorithm and its application
### 3.1 Framework
PPO is based on policy gradient optimization, with the core objective of maximizing cumulative rewards, which trains an intelligent agent to make decisions. The agent in this study can be seen as an AI decision-maker who decides whether to open a collection center and a warehouse for a humanitarian supply chain.

$$\max_{\theta} \mathbb{E}\left[\sum_{t=0}^{T} \gamma^t R_t\right] \quad (1)$$

where,

$\gamma$ is the discount factor, $\in (0, 1]$ is a discount factor.

$R_t$ is the reward received at time step $t$.

PPO uses a clipped surrogate objective function to update the policy while maintaining stability:

$$L^{\text{PPO}}(\theta) = \mathbb{E}_t\left[\min\left(\frac{\pi_\theta(a_t \mid s_t)}{\pi_{\theta_{\text{old}}}(a_t \mid s_t)} A_t, \text{clip}\left(\frac{\pi_\theta(a_t \mid s_t)}{\pi_{\theta_{\text{old}}}(a_t \mid s_t)}, 1-\epsilon, 1+\epsilon\right) A_t\right)\right] \quad (2)$$

where,

$\theta$ represents the policy network parameters,

$\pi_\theta(a_t \mid s_t)$ is the agent' probability of selecting action $a_t$ under the new policy, given the state of environment $s_t$ observed by it,

$\pi_{\theta_{\text{old}}}(a_t \mid s_t)$ is the probability of selecting $a_t$ under the old policy, given the state of environment $s_t$ observed by it,

$A_t$ is the advantage function (measuring how much better an action is compared to the expected action).

$\epsilon$ is the PPO clipping factor, to prevent overly large updates to the policy.

In which, the structure $s_t \rightarrow a_t \rightarrow s_{t+1}$ represents a Markov Decision Process (MDP).

### 3.2 Definition of HSC design problem

From the general HSC practices in the real world and academic papers, there should be collection centers for collecting donations from all social parties and delivering them to warehouses. Then, warehouses should distribute these kits to demand points where the impacted people gather.

This study aims to solve the efficient and equitable distribution at the same time for each time step, hence, this problem is not constrained by the time frame, but maximizes the supply delivered in each time step.

In this study, it is assumed that there are potential places within certain distance from the demand points that can be the locations of collection centers and warehouses, whether established or activated, and for each time step, like one day, the model can decide which collection centers or warehouses should be opened, keeping open or should be switched and calculate the delivering amount of donations to downstream for main considerations of efficiency and equity for HSCs in emergency.

Other assumptions are outlined as follows:

- Once one or several collection centers are activated at time $t$, their inventories equal their capacity, which are random integers from a uniform distribution within a value range.

- Once one or several collection centers are activated at time $t$, the donations they receive will be packed into uniform kits.
- When the inventory of a warehouse is out of stock after replenishment from collection centers, this warehouse is expected to outsource the necessities for kits.

All the parameters are listed in Appendix A and adjusted for the sensitivity test to illustrate different settings and situations. Since the lack of realistic time series demand data, various simulation methods are employed to mimic the volatility of demand.

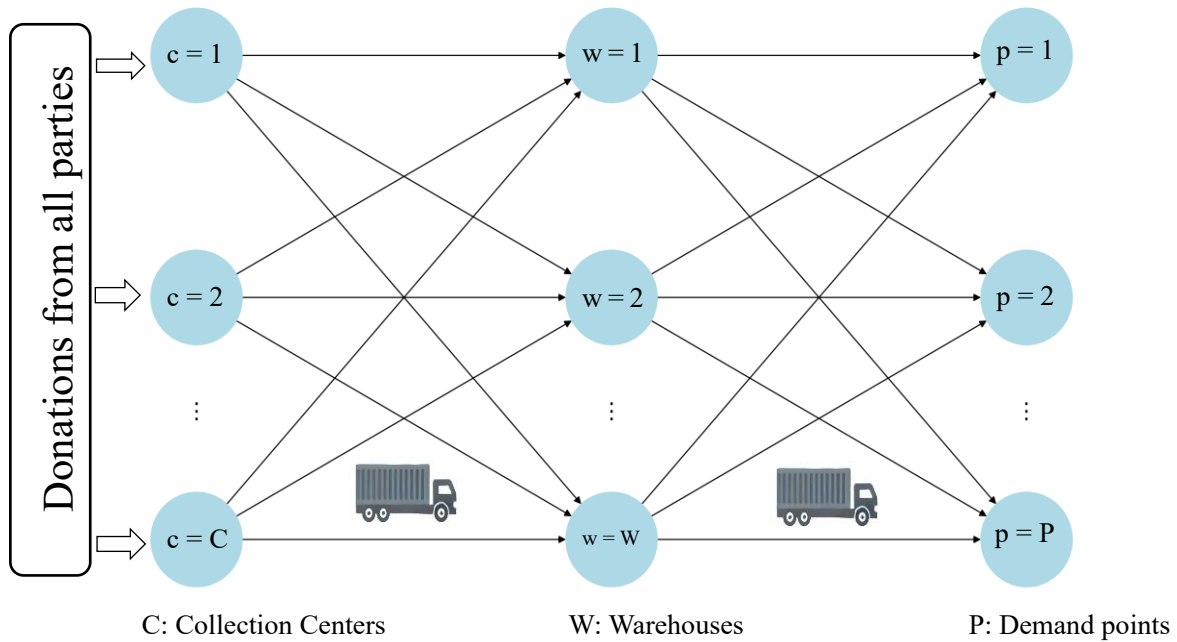

**Figure 1.** The network of designed HSC

### 3.3 The PPO optimisation model
All the sets, indices, notation, parameters, and decision variables are set as follows.

**Sets and indices:**
C: set of collection centers, indexed by $c$ ($c \in C$)
W: set of warehouses, indexed by $w$ ($w \in W$)
P: set of demand points, indexed by $p$ ($p \in P$)
T: set of time, indexed by $t$ ($t \in T$)

**Parameters:**
$D_{cw}$: the distance between the collection center $c$ and the warehouse $w$
$D_{wp}$: the distance between the warehouse $w$ and demand point $p$
$E_c$: the establishment cost of the collection center $c$
$E_w$: the establishment cost of warehouse $w$
$d_p^t$: the demand of demand point $p$ at time $t$

$C_c$: the capacity of collection center $c$
$C_w$: the capacity of warehouse $w$
$h$: the transportation cost per unit of kit and per unit of distance
$V$: the market value per unit of kit

**Decision variables:**
$x_c^t$: 1, if the collection center $c$ is established at time $t$, and 0, otherwise
$x_w^t$: 1, if the warehouse $w$ is established at time t, and 0, otherwise
$A_{cw}^t$: the amount of products distributed from the collection center $c$ to the warehouse $w$ at time $t$, the replenishment
$A_{wp}^t$: the amount of products distributed from warehouse $w$ to demand point $p$ at time $t$
$O_w^t$: the amount of outsourcing of the warehouse $w$ at time $t$

**State variables:**
$I_w^t$: Inventory level of warehouse $w$ at time $t$
$u_p^t$: the number of unsatisfied demands for demand point $p$ at time $t$
$r_p^t = \frac{A_{wp}^t}{d_p^t}$: Satisfaction rate of demand point $p$ at time $t$

$\bar{r}^t = \frac{1}{|P|}\sum_{p \in P} r_p^t$: the average satisfaction rate

$\bar{I}^t = \frac{1}{|W|}\sum_{w \in W} I_w^t$: the average inventory of each warehouse $w$ at time $t$

**Reward Function**
The PPO agent's primary objective is to minimize costs while satisfying demand efficiently, using the cost-efficiency format

$$R_t = \frac{f_{\text{efficiency}}(t)}{f_{\text{cost}}(t)} \tag{3}$$

Then, take the log transformation to facilitate the algorithm learning

$$R_t = log\left(\frac{1 + f_{\text{efficiency}}(t)}{1 + f_{\text{cost}}(t)}\right) = \log(1 + f_{\text{efficiency}}(t)) - log(1 + f_{\text{cost}}(t)) \tag{4}$$

where,

$$f_{\text{efficiency}}(t) = \sum_{p \in P}(d_p^t - u_p^t) \times V \tag{5}$$

this study monetarized the efficiency function for direct comparison,

$$f_{\text{cost}}(t) = \sum_{c \in C} E_c x_c^t + \sum_{w \in W} E_w x_w^t + h \left( \sum_{c \in C} \sum_{w \in W} D_{cw} A_{cw}^t + \sum_{w \in W} \sum_{p \in P} D_{wp} A_{wp}^t \right) + Penalty \tag{6}$$

$$Penalty = Penalty_{\text{mismatch}} + Penalty_{\text{switch}} \tag{7}$$

$$Penalty_{\text{mismatch}} = \lambda_{penalty} \sum_{j=1, j \in w}^{W} \left| \sum_{i=1, i \in c}^{C} A_{cw}^t(i,j) - \sum_{k=1, k \in p}^{P} A_{wp}^t(j,k) \right| \tag{8}$$

$$Penalty_{\text{switch}} = \lambda_{Switch\ Penalty_c} \sum_{i=1, i \in c}^{C} |x_c^t - x_c^{t-1}| + \lambda_{Switch\ Penalty_c} \sum_{k=1, k \in p}^{P} |x_w^t - x_w^{t-1}| \tag{9}$$

Other constraints as follows,

$$\sum_{w \in W} A_{cw}^t \leq C_c x_c^t; \forall c \in C \text{ and } \forall t \in T \tag{10}$$

$$\sum_{p \in P} A_{wp}^t \leq I_w^t x_w^t; \forall w \in W \text{ and } \forall s \in T \tag{11}$$

$$A_{cw}^t \geq 0 \text{ and integer}; \forall c \in C, \forall w \in W, \text{ and } \forall t \in T \tag{12}$$

$$A_{wp}^t \geq 0 \text{ and integer}; \forall w \in W, \forall p \in P, \text{ and } \forall t \in T \tag{13}$$

$$x_c^t \in \{0,1\}; \forall c \in C \tag{14}$$

$$x_w^t \in \{0,1\}; \forall w \in W \tag{15}$$

**The state and action space**

This study uses Partially Observable Markov Decision Process (POMDP) to let the state $s_t$ gained from the observation consists of current demand $d_p^t$, current warehouse inventor $I_w^t$, previous levels $A_{wp}^{t-1}$, and the current time step $t$, as $s_t = \{d_p^t, I_w^t, A_{wp}^{t-1}, t\}$. They are key observations that enable the agent to learn dynamic supply chain strategies by integrating historical and real-time information. This structured representation allows the model to optimize resource allocation while adapting to fluctuating demand and supply constraints. Action Space for the PPO agent is to decide the opening, closing, or to put it another way, switching of collection centers and warehouses as $A_t = \{x_c^t, x_w^t\}$.

**Computation of the distributions and inventory**

The Computation of the collection center to the warehouse distribution $A_{cw}^t$:

$$A_{cw}^t = (\text{Dirichlet Distribution}) \times C_c \times x_c^t \tag{16}$$

It uses Dirichlet allocation to distribute kits from collection centers to warehouses, and only active collection centers $(x_c^t = 1)$ can participate in the distribution.

Firstly, the computation of kits from activated warehouses to demand point distribution $A_{wp}^t$ is randomly sampled from a discrete uniform distribution.

$$A_{wp}^t \sim \text{UniformInt}(0, \max(I_w^t \cdot x_w^t, 1)) \tag{17}$$

where, $I_w^{t+1} = I_w^t + \sum_{c \in C} A_{cw}^t - \sum_{p \in P} A_{wp}^t + O_w^t$ (18), and $I_w^t \geq 0$ (19). The warehouse inventory level ($I_w^t$) determines how many kits can be shipped and only active warehouses ($x_w^t = 1$) can participate in the distribution. If an activated warehouse is out of stock, then this warehouse will outsource kits by itself, and the penalty of mismatching will be its costs. Otherwise, the mismatching penalty is the cost of holding. This study also considers the situation of including the $I_w^{t-1}$, which gives $Penalty_{\text{dismatch}} = \lambda_{penalty} \sum_{j=1, j \in W}^{W} |I_w^{t-1} + \sum_{i=1, i \in c}^{C} A_{cw}^t(i,j) - \sum_{k=1, k \in p}^{P} A_{wp}^t(j,k)|$. The results are shown in Appendix B.13. Through these, undersupply can be excluded. The $\max(\cdot, 1)$ ensures the validity of the sampling range even when the available inventory is zero.

Then,

$$A_{wp}^t = \left\lfloor A_{wp}^t \cdot \min\left(1, \frac{d_p^t}{\sum_{w \in W} A_{wp}^t + \varepsilon}\right) \right\rfloor \tag{20}$$

$A_{wp}^t$ is adjusted for demand scale, because there will be more than one warehouse delivering the kits to one demand point. The demand level of the demand point $p$ ($d_p^t$) determines the maximum required number of distributions. That is to prevent both undersupply and oversupply.

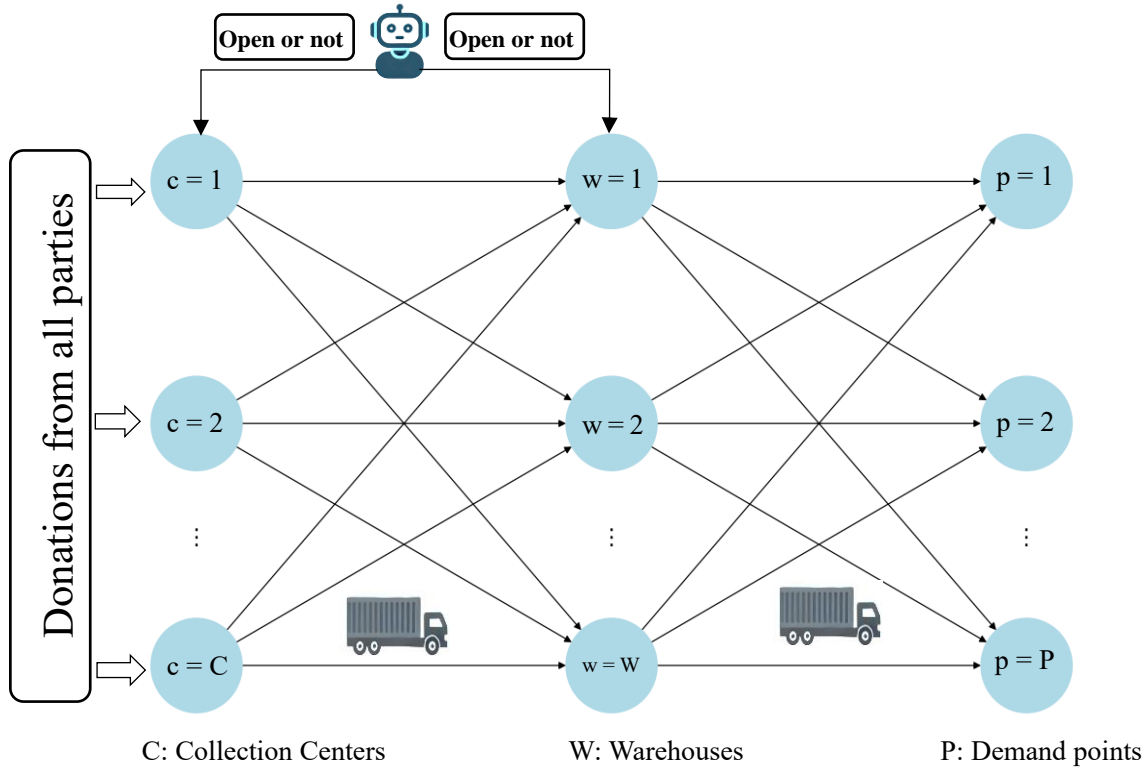

**Figure 2.** The network of HSC with AI agent of PPO

### 3.4 Demand Initialization
To simulate a volatile environment where dynamic demand of each demand point is likely to happen, this study generates it through three simulation ways for stochastic and deterministic data. The seed is of 42 when it is applicable.

The equation of GBM Demand is $d_p^t = d_p^{t-1} \times exp\left(\left(\mu - \frac{1}{2}\sigma^2\right) + \sigma dW + \text{shock}\right)$, where, $\mu =$ Drift term (0.02), $\sigma =$ Volatility (0.1), $dW =$ Standard Brownian Motion increment, Shock = Random disturbance.

The equation of Poisson Demand is updated as $d_p^t = d_p^{t-1} + \Delta_t$, where, $\Delta_t \sim \text{Poisson}(\theta)$, which is random demand increment at time $t$, $\theta$ is the expected value of the Poisson distribution controlling demand increase (30), and $d_{p,0} \sim \text{Uniform}(1200, 2000)$.

The equation of Merton Demand (Merton, 1976) is evolved as $d_p^t = d_p^{t-1} \cdot exp\left(\left(\mu - \frac{1}{2}\sigma^2\right) + \sigma \cdot \varepsilon_t + J_t\right)$, where, $\mu =$ the average drift or growth rate (0.02), $\sigma =$ the volatility of normal fluctuations (0.1), $\varepsilon_t \sim \mathcal{N}(0,1)$: standard normal random noise, $\theta$: the jump intensity, or, the rate of jump events (0.1), $J_t$: Jump Size $\sim \mathcal{N}(\text{jump\_mean}, \text{jump\_std}^2)$, jump size drawn from a normal distribution with mean 0.05 and standard deviation 0.2, if $\text{Poisson}(\theta) > 0$, else 0.

The following are examples of stochastic data produced by the GBM and Merton simulation methods, and the one of Poisson is shown in the Appendix B.17.

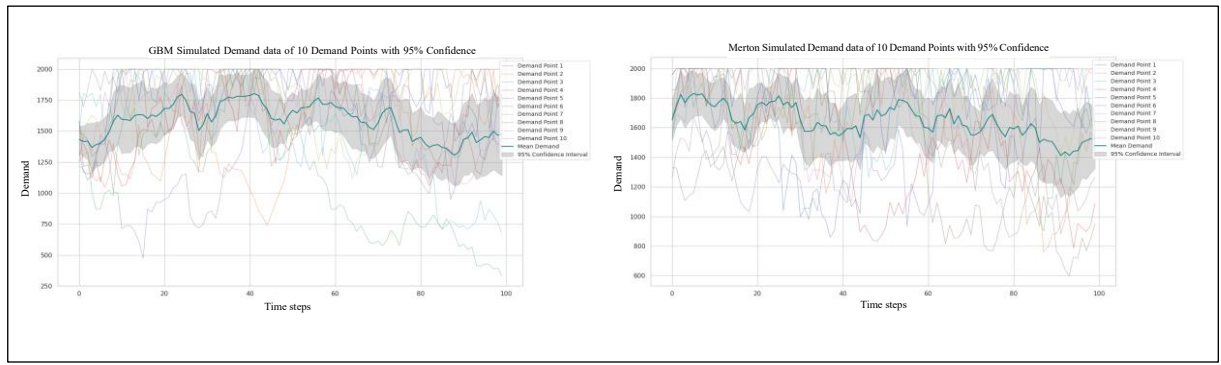

**Figure 3.** Examples of different demands and confidence intervals

# 4 Numerical Results

## 4.1 The result from the PPO algorithm

The following is the results from PPO with stochastic GBM demands, which are based on the table of parameters in Appendix A and displayed in the form of average value over 5 runs per episode. Before perring episode across 4 runs, the average satisfaction rate is originally referred to $\bar{r}^t$, which is averaged over each demand point at each time step, and the original average inventory is $\bar{I}^t$, which is the average inventory for each warehouse at each time step. The rest metrics are the values of functions of reward ($R_t$), efficiency ($f_{\text{efficiency}}(t)$), cost ($f_{\text{cost}}(t)$). The ultimate goal is to guide the average satisfaction rate to reach the optimal level, which is 1, representing the fully fulfilled efficient and equitable distribution in HSC. This goal is embedded in the objective functions. Through optimizing the reward function, the agent trained by the PPO will raise the average satisfaction based on the observations, objectives, and its own judgment.

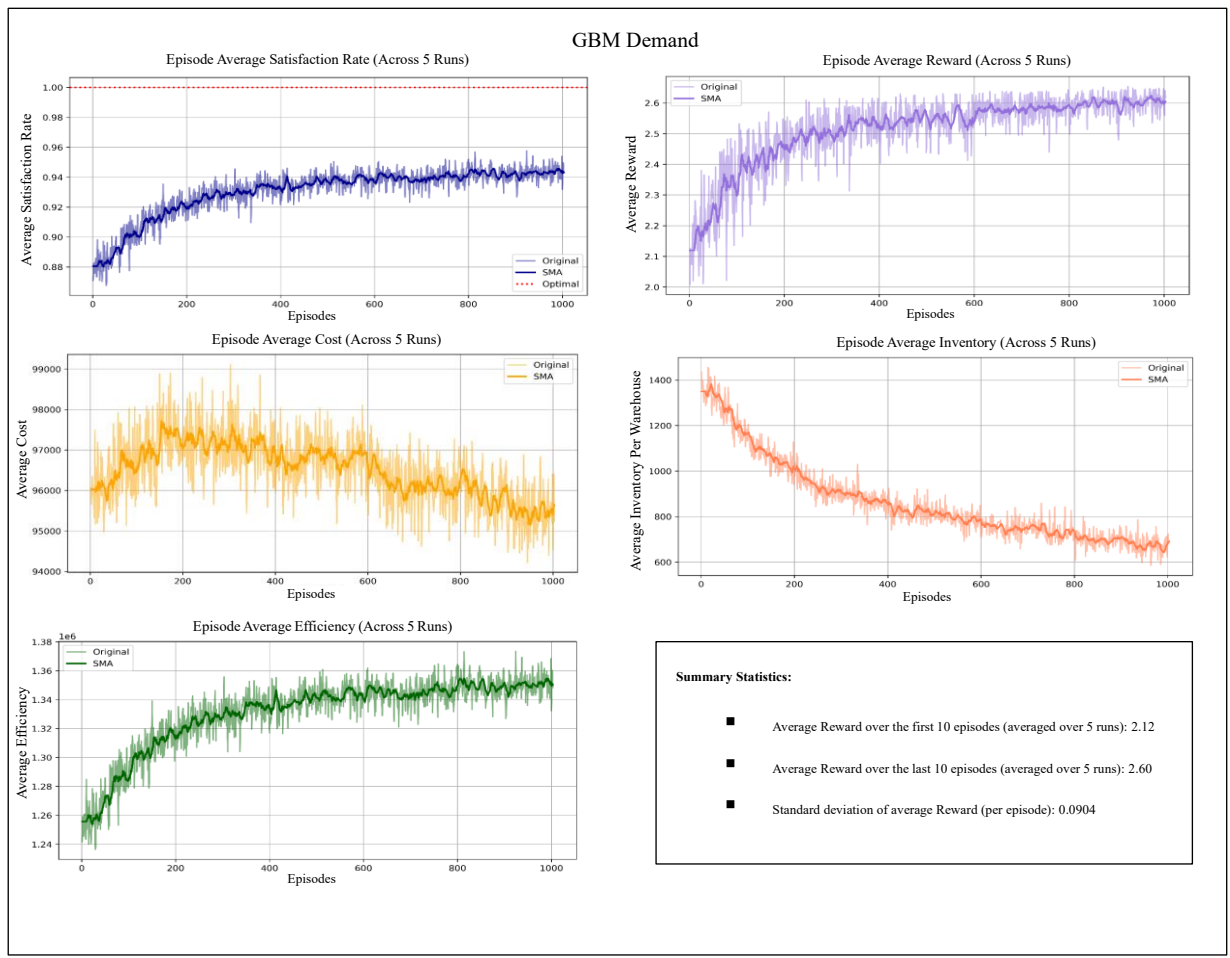

**Figure 4.** Results after Training Test in HSC with GBM stochastic demands.

From the above chart, it can be seen that the evident upward trends for average satisfaction rate, reward and efficiency, and same obvious downward trends in cost and average inventory. These are ideal results. However, the trends of costs upwards at different degrees on stochastic Poisson and Merton demand data with other trends of resting metrics remaining same.

Figure 5 below shows key metrics as average satisfaction rate, which represents the efficiency and equity in this designed HSC, and reward, which represents the model's learning performance with Poisson and Merton demands. Other metrics with these two demand data are attached in Appendix B.1 and B.2.

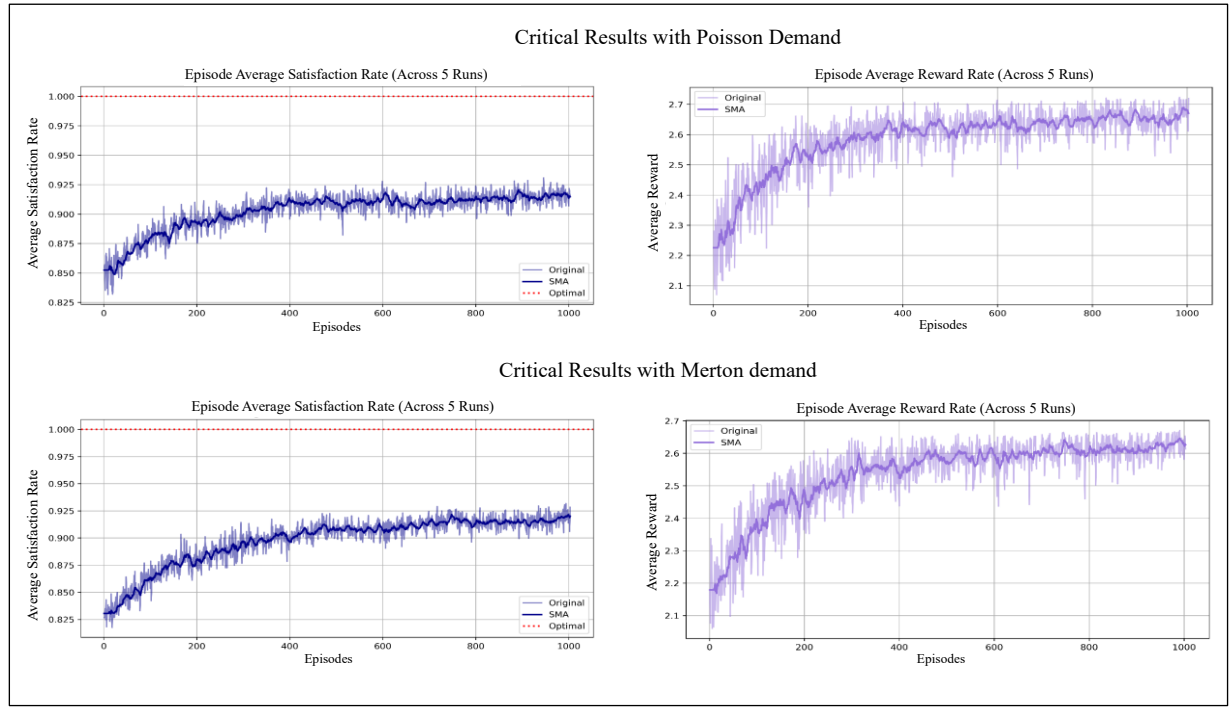

**Figure 5.** Critical Results after Training Test in HSC with stochastic Poisson and Merton demands.

**Table 1.** The summary of average reward (Avg Reward) of the first and last 10 episodes and standard deviation (Std) across 5 runs with different demands

| Demands | | Results from 5 Runs | | |
|---|---|---|---|---|
| Name | Clipped Value Range | Avg Reward (First 10 Episodes) | Avg Reward (Last 10 Episodes) | Reward Std |
| GBM | 0-2000 | 2.12 | 2.60 | 0.0904 |
|  | 0-1000 | 1.78 | 2.33 | 0.0926 |
|  | 0-3000 | 2.18 | 2.67 | 0.0898 |
| Poisson | 0-2000 | 2.23 | 2.67 | 0.0965 |
|  | 0-1000 | 2.03 | 2.47 | 0.0989 |
|  | 0-3000 | 2.23 | 2.68 | 0.0867 |
| Merton | 0-2000 | 2.18 | 2.63 | 0.0891 |
|  | 0-1000 | 1.96 | 2.41 | 0.0949 |
|  | 0-3000 | 2.21 | 2.66 | 0.0886 |

Table 1 summarizes that the first 10 episodes' rewards are consistently higher than those of the last 10

episodes and proves that the PPO works in this HSC under different simulated demand scenarios for different clipping values of demands the agent can observe. The upper bound of 2000 is an initialized parameter of the environment setting, which is shown in Figure 4 and discussed. For the upper bound of 1000, all the costs of the three simulated methods are decisively downwards, while, in the case of 3000, all the costs appear upwards for satisfying the drastic increase of demand. The other four metrics remain the same trends in these different situations. The corresponding results are provided in Appendices B.3-B.7.

**4.2 The sensitivity test from PPO**

Results of PPO of geometric Brownian motion (GBM) under different parameter settings with deterministic GBM demand data. This test can be seen as the fluctuating or interrupting environment that impacts on the potential numbers of facilities, the capacity of facilities, the distance range, cost of outsourcing and switching, when the demand of each demand points is deterministic.

**Table 2.** The comparison of rewards of the first and last 10 episodes and the standard deviation across 3 runs with different parameter settings

| Parameters | | Results from 3 Runs with deterministic demand data | | |
|---|---|---|---|---|
| Name | Value | Avg Reward (First 10 Episodes) | Avg Reward (Last 10 Episodes) | Reward Std |
| Number of collection centers; Number of warehouses; Number of demand points. | ① 15; 5;10 | 2.03 | 2.57 | 0.0772 |
| | ② 5;10;15 | 2.85 | 2.85 | 0.0369 |
| | ③ 15,10, 5 | 2.19 | 2.26 | 0.0209 |
| Distances. | ④ 5-100 | 0.40 | 0.84 | 0.0521 |
| Activating cost of collection center; Capacity of collection center; | ⑤ 500-1000 600-1125 300-700 2000-5000 | 2.22 | 2.63 | 0.0741 |
| Activating cost of warehouse; Capacity of warehouse. | ⑥ 300-700 2000-5000 500-1000 600-1125 | 1.46 | 2.18 | 0.0677 |
| Coefficient of mismatching; | ⑦ 0.5, 15, 20 | 2.25 | 2.64 | 0.0840 |
| Coefficient of switching collection center; | ⑧ 10, 5, 3 | 1.27 | 1.97 | 0.0663 |
| Coefficient of switching warehouse | ⑨ 5, 50, 25 | 1.82 | 2.26 | 0.0712 |

From Table 2, overall, the model can learn, no matter how the parameters change. The worst case is ② when 5 collection center, 10 warehouses and 15 demand points that the average reward of last 10 episodes is equal as that the first 10 episode since that combinations means the original supply represented by the collection centers is far short so that the model cannot improve the current strategy to achieve higher reward.

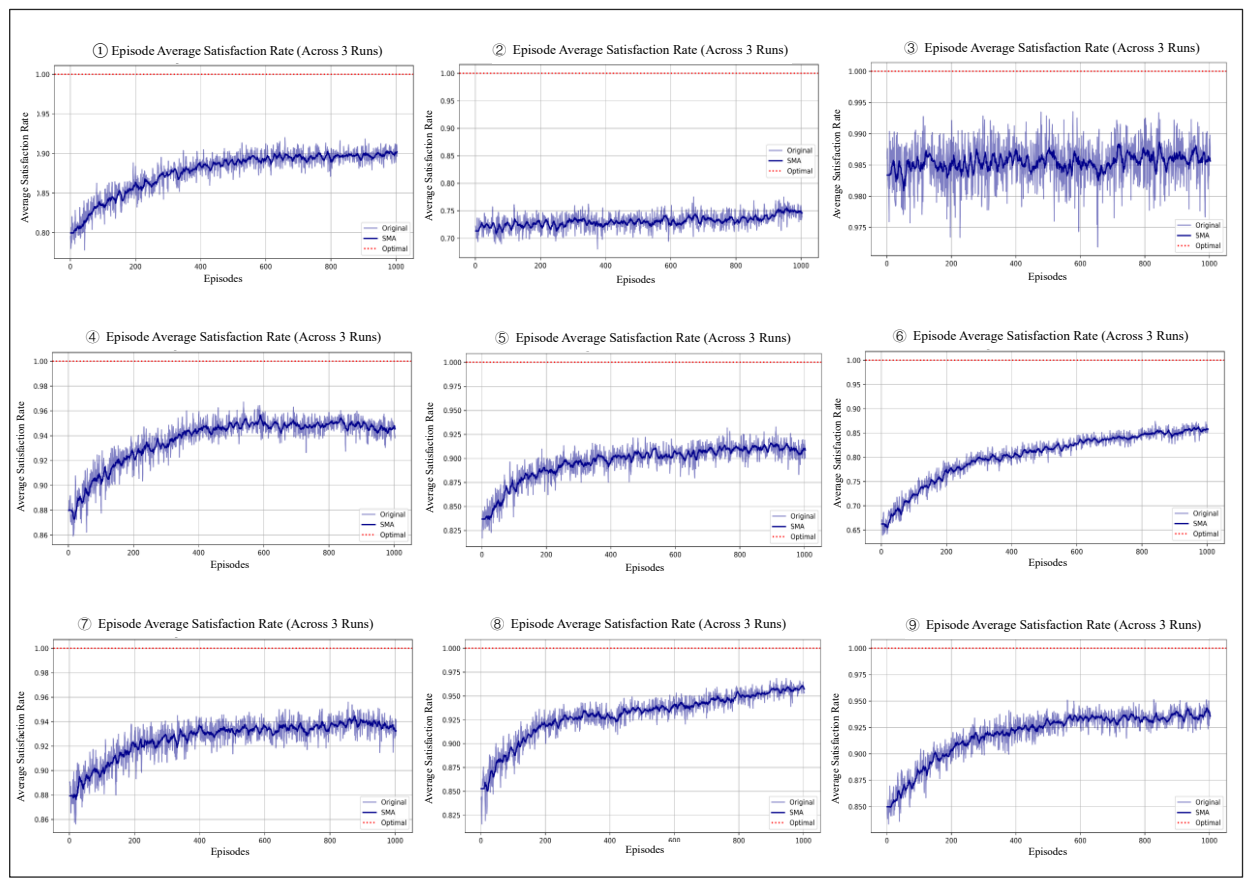

**Figure 6.** The comparison of the average satisfaction rate under different parameter settings

All the trends of the average satisfaction rate have increased to varying degrees. Case ② and case ③ are least obvious. Case ② shows the lowest average satisfaction rate, which is roughly below 0.75 all the way because of too few potential collection center can supply the kits, and in case ③, which is generally at 0.985, nearly saturated to 1 with no space to improve, along with the 100 time steps because the number of collection centers exceed the numbers of warehouses and demand points.

In this place, only the average satisfaction rate is displayed because it is the most paramount goal after proving the PPO works in Table 1. Other results of the rest metrics can be found in the Appendix B.8-B.11. Except for the average reward discussed, the average efficiency grows for different setting as well. Whereas, the trends of average cost in Case ①, ②, ⑤ and ⑦ go upwards, and the trends of average inventory of case ② and ③ are most fluctuate, in which, case ② even shows the average inventory rises again after a slight decline within 220 and 320 approximately since only 5 potential collection center in this case making the model try best to meet the each point's demand and sacrifices the downwards pattern that presents the effective inventory management.

The following Figure 7 is the sensitive test with two prices, one is the unit transportation cost $h$, and the other is the market value $V$ of one kit. The Figure 7 lists the results for parameter settings from ① to ⑥.

The rest of the results are included in Appendix B.12.

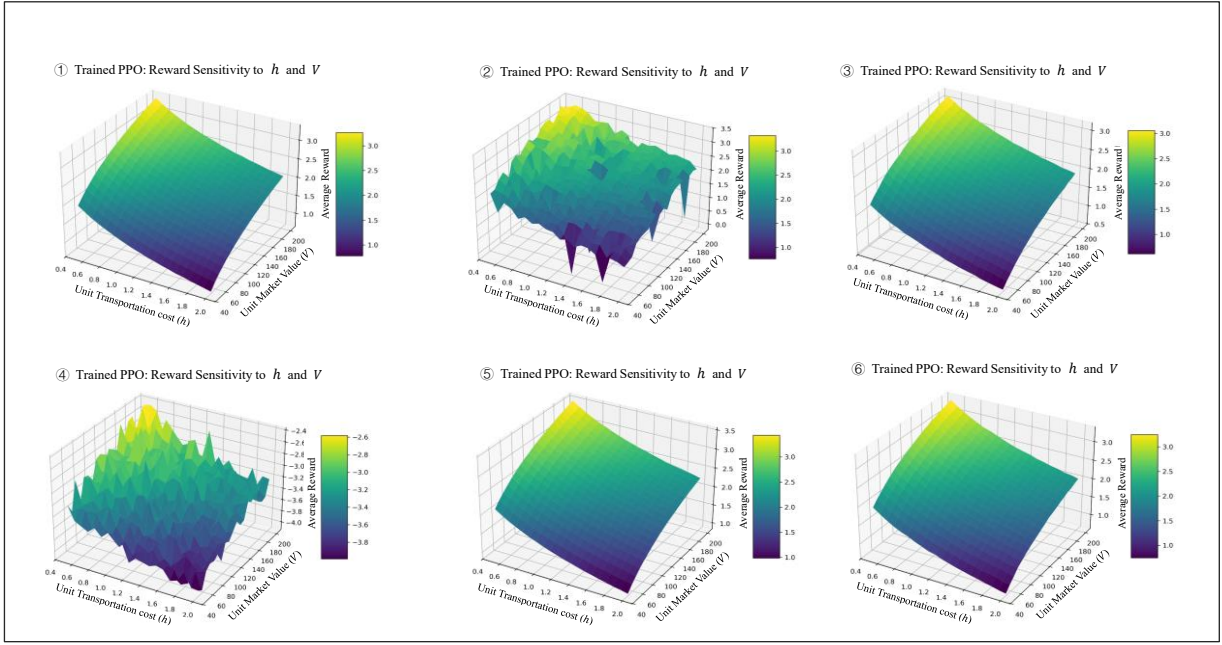

**Figure 7.** The comparison of the effects of $h$ and $V$ on average reward after training under different parameter settings

All the results are smooth except under the cases of ② and ④, since the lack of the original supply in case ② and significantly increasing the range of the distance in case ④ make the average reward especially sensitive to $h$ and $V$, correspondingly, as equation (3) states. In the training process of Case ②, the agent improves the average satisfaction rate within a limited range because of the constraints of the initial supply, just as the situation could happen in reality, hence underscoring $V$ when chasing a high value of reward and causing instability as shown. In the Case ④, it exaggerates the cost effect since the enlarged distance makes the effect of $h$ significant.

### 4.3 Comparisons with heuristic algorithm

The final comparison is conducted with other heuristic methods. Non-dominated Sorting Genetic Algorithm II (NSGA-II) is a popular and efficient multi-objective optimization algorithm, introduced by Deb et al. in 2002. It is widely used to solve problems with conflicting objectives. This study applies NSGA-II by using the balance score (BS), which is a trade-off approach and best efficiency (BE) index approach that adopts the strategy under the best efficiency situation, to choose the solution from the Pareto front, respectively, maximizing the efficiency and minimizing the cost simultaneously. Particle swarm optimization (PSO) is a global optimization algorithm based on swarm intelligence, proposed by Kennedy and Eberhart in 1995. It is inspired by the foraging behavior of bird flocks. This study uses a single-objective application of PSO as the PPO.

The experiment setting involves running PPO three times, evaluating it once, and then running the other three heuristic methods, followed by evaluation of each method once.

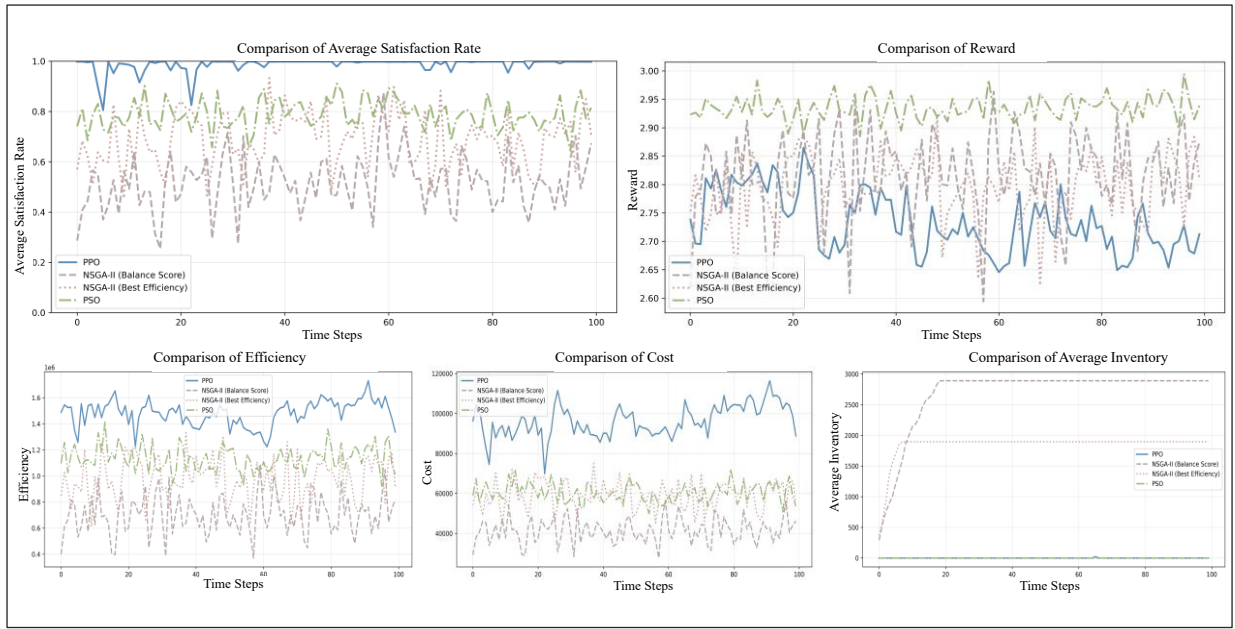

**Figure 8.** The comparison of heuristic methods on five metrics with GBM demand data under different parameter settings

The curve charts above show the comparisons between four methods by five metrics over 100 time steps. It can be seen that the average satisfaction rate of PPO is above other three method and more stable over the later time stage, even the reward goes down slightly, implicating that it doesn't choose to reduce the supply through compromised with the cost, but opens and switches the collection centers and warehouses frequently and delivers kits efficiently and equitably to meet the higher satisfaction rate and efficiency. Thereby, the cost contains the activating, switching, and delivering cost increase as well. PPO and PSO are better off than two NSGA-II approaches in inventory management, while best efficiency approach is better in improving average satisfactory rate than the other one. Correspondingly, the cost of it is higher than the balance score as well.

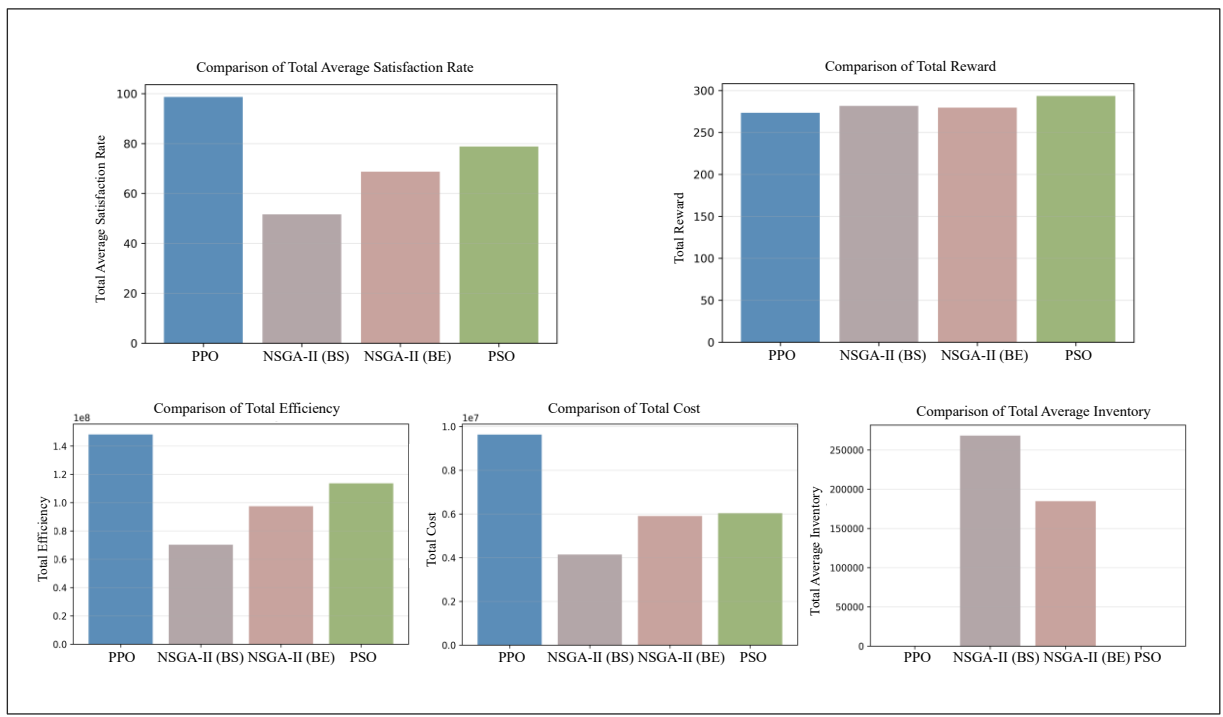

**Figure 9.** The comparison of heuristic methods on total values of five metrics with GBM demand data under different parameter settings

From these total value of the five metrics from the 100 time steps, it can be clearly seen the difference between these method. Both multi-objective approaches of NSGA-II compromise much on average satisfaction rates for the sake of cost to achieve a minute higher value of total reward than PPO, and are more poorer performance in inventory management compared with PPO. While, since PSO uses single objective approach, it only focuses on maximizing the reward with 20% in average satisfaction rate and efficiency than PPO.

The corresponding results of five metrics with Merton and Poisson demands are attached in the Appendices B.14 and B.15. The average satisfaction rate and efficiency of PPO are superior to the other threes methods with these two different demand data as well.

### 4.3.2 Discussion

This study fully answers the research questions by combining the efficiency and equity together through model design and considering the cost, but not only focusing on the cost.

For RQs-1, the reward function as the explicit objective guiding the agent maximizing the efficiency through expanding distribution and minimizing the costs related to transportation, activate and switch the facilities, for the purpose of improving the efficiency. In the meantime, the number of kits delivered from each activated warehouse to every demand point is constrained by the inventory level, outsourcing penalty and scaled by the proportion to meet the each demands, thus preventing undersupply and oversupply for the purpose of equity to realize the implicit objective of raising average satisfaction rate. Therefore, the efficiency, equity and cost are infused through the logic of the model designed.

For RQs-2, this study uses different demand data from GBM, Poisson and Merton to simulate more undulant and volatile than realistic situation as examples of optimizing the design of HSC. With setting parameters are unchanged, reward consistently grows for different data pattern regardless of how the trends of cost have been drawn by the dramatic increasing and decreasing of the upper bound of the demand data. Meanwhile, all the outcomes display upward trends in average satisfaction rate and efficiency. That demonstrates the adaptability of PPO as one of RL algorithm for optimization in dynamically varied demand data and illustrates the process of adapting clearly during training over time steps.

In term of RQs-3, although changing parameter setting impacts the level of average satisfaction rate and the performance of reward, the uptrends of them and efficiency are remained, and most inventory shows downtrends, the cost varies as usual, indicating the model works in different situations, like the decrease of initial supply when number of collection center declines, the decrease of ability to store and transport the kit when number of warehouse declines, the longer distance when normal routines collapse or varied by the secondary disasters or other emergencies, the activating costs and capacities of collection centers and warehouses when affected by the changing environment, the cost of holding and outsourcing kits as well as switching facilities during crisis. All of these implicates the model based on the PPO pay close attention on improving the average satisfaction rate, the ultimate goal, throughout.

Towards RQs-4, the average satisfactory rate of PPO surpasses the other three methods when comparing horizontally since the AI agent are trained to observe the environment as stated of $s_t$ and it learns the average satisfaction rate for HSC is the priority, even pulling the cost higher to meet the needs of demand points, which is wiser of taking the cost as the second consideration when maximizing the reward function. PPO is superior at raising the average satisfaction rate, not only pursuing the higher reward.

## 5. Conclusion

To sum up, this HSC design under the PPO are proved that it can deal with many various situations when emergency happens. There are great potential in RL like PPO to develop for instant response and getting rid of bias.

There are some limitation exists in this study, since lacking the realistic time-serious demand data, this study uses simulated method to mimic real world data as most other paper in HSC did. Although this is conducted between traditional single-objective and multi-objective heuristic algorithms, but not compared with other reinforcement learning methods. This study assumes delivery happening for every time step, not considering the delay effects may happened in real world.

If such algorithm can be integrated with remote sensing technology which detect and record the times-series data, and cooperate with other reinforcement learning algorithms to improve the efficiency and equity in HSC in reality.

## Reference

Abushaega, M. M., González, A. D., Moshebah, O. Y., 2024. A fairness-based multi-objective distribution and restoration model for enhanced resilience of supply chain transportation networks. Reliability Engineering & System Safety 251, 110314.

Cheng, J., Feng, X., Bai, X., 2021. Modeling equitable and effective distribution problem in humanitarian relief logistics by robust goal programming. Computers & Industrial Engineering 155, 107183.

Deb, K., Pratap, A., Agarwal, S., Meyarivan, T., 2002. A fast and elitist multiobjective genetic algorithm: NSGA-II. IEEE Transactions on Evolutionary Computation 6 (2), 182–197.

Dupuit, J., 1952. On the measurement of the utility of public works (Barback, R. H., Trans.). International Economic Papers 2, 83–110. (Orig. work pub. 1844)

Erokhin, D., 2025. Public discourse surrounding the 2025 California wildfires: A sentiment and topic analysis of high-engagement YouTube comments. Geosciences 15 (3), 100.

Ershadi, M.M., Shemirani, H.S., 2022. A multi-objective optimization model for logistic planning in the crisis response phase. Journal of Humanitarian Logistics and Supply Chain Management 12 (1), 30–53.

Haavisto, I., Kovács, G., 2015. A framework for cascading innovation upstream the humanitarian supply chain through procurement processes. Procedia Engineering 107, 140–145.

Kaur, H., Singh, S.P., 2022. Disaster resilient proactive and reactive procurement models for humanitarian supply chain. Production Planning & Control 33 (6–7), 576–589.

Kennedy, J., Eberhart, R., 1995. Particle swarm optimization. In IEEE Int. Conf. Neural Netw. 4, 1942–1948.

Khalili-Fard, A., Hashemi, M., Bakhshi, A., Yazdani, M., Jolai, F., Aghsami, A., 2024. Integrated relief pre-positioning and procurement planning considering NGO support and perishable relief items in a humanitarian supply chain network. Omega 127, 103111.

Khorsi, M., Chaharsooghi, S. K., Bozorgi-Amiri, A., Hussein-zadeh Kashan, A., 2020. A multi-objective multi-period model for humanitarian relief logistics with split delivery and multiple uses of vehicles. Journal of Systems Science and Systems Engineering 29 (3), 360–378.

Liu, X., Hu, M., Peng, Y., Yang, Y., 2024. Multi-Agent Deep Reinforcement Learning for Multi-Echelon Inventory Management. Production and Operations Management 1–21.

Maghsoudi, A., Harpring, R., Piotrowicz, W. D., Heaslip, G., 2021. Cash and voucher assistance along humanitarian supply chains: A literature review and directions for future research. Disasters 45 (4), 741–769.
Modarresi, S. A., Maleki, M. R., 2023. Integrating pre and post-disaster activities for designing an equitable humanitarian relief supply chain. Computers & Industrial Engineering 181, 109342.

Noham, R., Tzur, M., 2018. Designing humanitarian supply chains by incorporating actual post-disaster decisions. European Journal of Operational Research 265 (3), 1064–1077.

Puett, C., 2019. Assessing the cost-effectiveness of interventions within a humanitarian organisation. Disasters 43 (1), 575–590.

Schulman, J., Levine, S., Abbeel, P., Jordan, M., Moritz, P., 2015. Trust Region Policy Optimization. In 32nd Int. Conf. Mach. Learn. (ICML), 1889–1897.

Schulman, J., Wolski, F., Dhariwal, P., Radford, A., Klimov, O., 2017. Proximal Policy Optimization Algorithms. arXiv preprint arXiv:1707.06347.

Timperio, G., Kundu, T., Klumpp, M., de Souza, R., Loh, X. H., Goh, K., 2022. Beneficiary-centric decision support framework for enhanced resource coordination in humanitarian logistics: A case study from ASEAN. Transportation Research Part E: Logistics and Transportation Review.

Torabi, S.A., Shokr, I., Tofighi, S., Heydari, J., 2018. Integrated relief pre-positioning and procurement planning in humanitarian supply chains. Transportation Research Part E: Logistics and Transportation Review 113, 123–146.

Van Steenbergen, R., Mes, M., Van Heeswijk, W., 2023. Reinforcement learning for humanitarian relief distribution with trucks and UAVs under travel time uncertainty. Transportation Research Part C: Emerging

Technologies 157, 104401.

Wang, Z., Leng, L., Ding, J., Zhao, Y., 2023. Study on location-allocation problem and algorithm for emergency supplies considering timeliness and fairness. Computers & Industrial Engineering 177, 109078.

Wiener, J. B., 2013. The diffusion of regulatory oversight. In: Livermore, M. A., Revesz, R. L. (Eds.), The globalization of cost–benefit analysis in environmental policy. Oxford Univ. Press, pp. 123–144.

Yu, L., Zhang, C., Jiang, J., Yang, H., Shang, H., 2021. Reinforcement learning approach for resource allocation in humanitarian logistics. Expert Systems with Applications 173, 114663.

Yılmaz, Ö. F., Guan, Y., Gürsoy Yılmaz, B., 2025. Designing a resilient humanitarian supply chain by considering viability under uncertainty: A machine learning embedded approach. Transportation Research Part E: Logistics and Transportation Review 194, 103943.

**Appendix**

A   The initialized parameters table

| Category | Description | Initialization Value |
| --- | --- | --- |
| Environment Settings | Number of collection centers | 15 |
| Environment Settings | Number of warehouses | 5 |
| Environment Settings | Number of demand points | 10 |
| Environment Settings | Number of time steps | 100 |
| Collection Centers | Setup cost for collection centers | 400-800, |
| Collection Centers | Capacity of collection centers | 800-1500 |
| Warehouses | Setup cost for warehouses | 200-500 |
| Warehouses | Capacity of warehouses | 4000-10000 |
| Transportation Costs | Transportation cost coefficient | 0.5 |
| Transportation Costs | Cost matrix: Collection centers → Warehouses | 5-10 |
| Transportation Costs | Cost matrix: Warehouses → Demand points | 5-10 |
| Demand Simulation | Mean demand growth rate | 0.02 |
| Demand Simulation | Demand volatility | 0.1 |

| Demand Simulation | Initial demand for each demand point | 1200-2000, Clip (0, 2000) |
|---|---|---|
| Reinforcement Learning | Discount factor for future rewards | 0.999 |
| Reinforcement Learning | Learning rate for PPO | 0.0001 |
| Reinforcement Learning | Entropy coefficient (exploration) | 0.01 |
| Reinforcement Learning | Training batch size | 64 |
| Reinforcement Learning | Total training steps | 20000 |
| Inventory Management | Initial warehouse inventory | 0 |
| Penalty & Cost Factors | Cost of mismatching kits the warehouse received and delivered | 1 |
| Penalty & Cost Factors | Cost for switching collection centers | 30 |
| Penalty & Cost Factors | Cost for switching warehouses | 10 |
| Efficiency Calculation | Market value of Per Kit | 100 |
| NSGA-II (BS&BE) | Population size per generation of NSGA-II | 200 solutions per generation |
| NSGA-II (BS&BE) | Number of generations of NSGA-II | 200 |
| PSO | Population size per generation of PSO | 100 |
| PSO | Number of generations of PSO | 200 |

B Results

B.1 Other metrics of Poisson demand (0-2000)

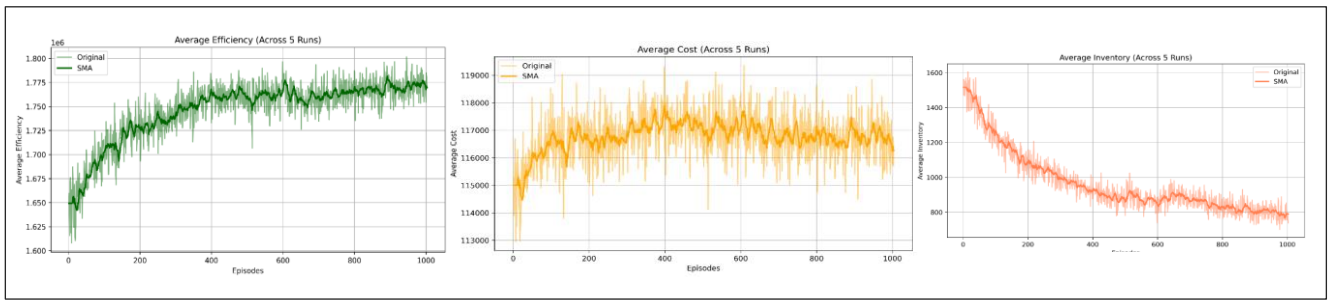

B.2 Other metrics of Merton demand (0-2000)

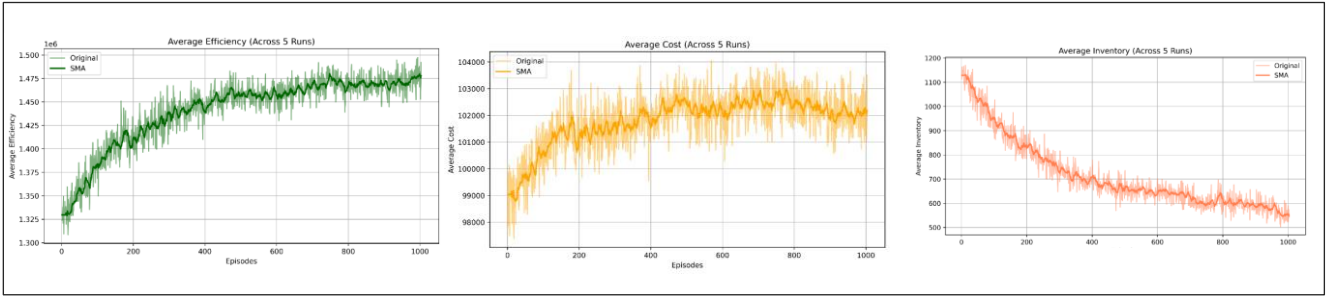

### B.3 GBM demand (0-3000)

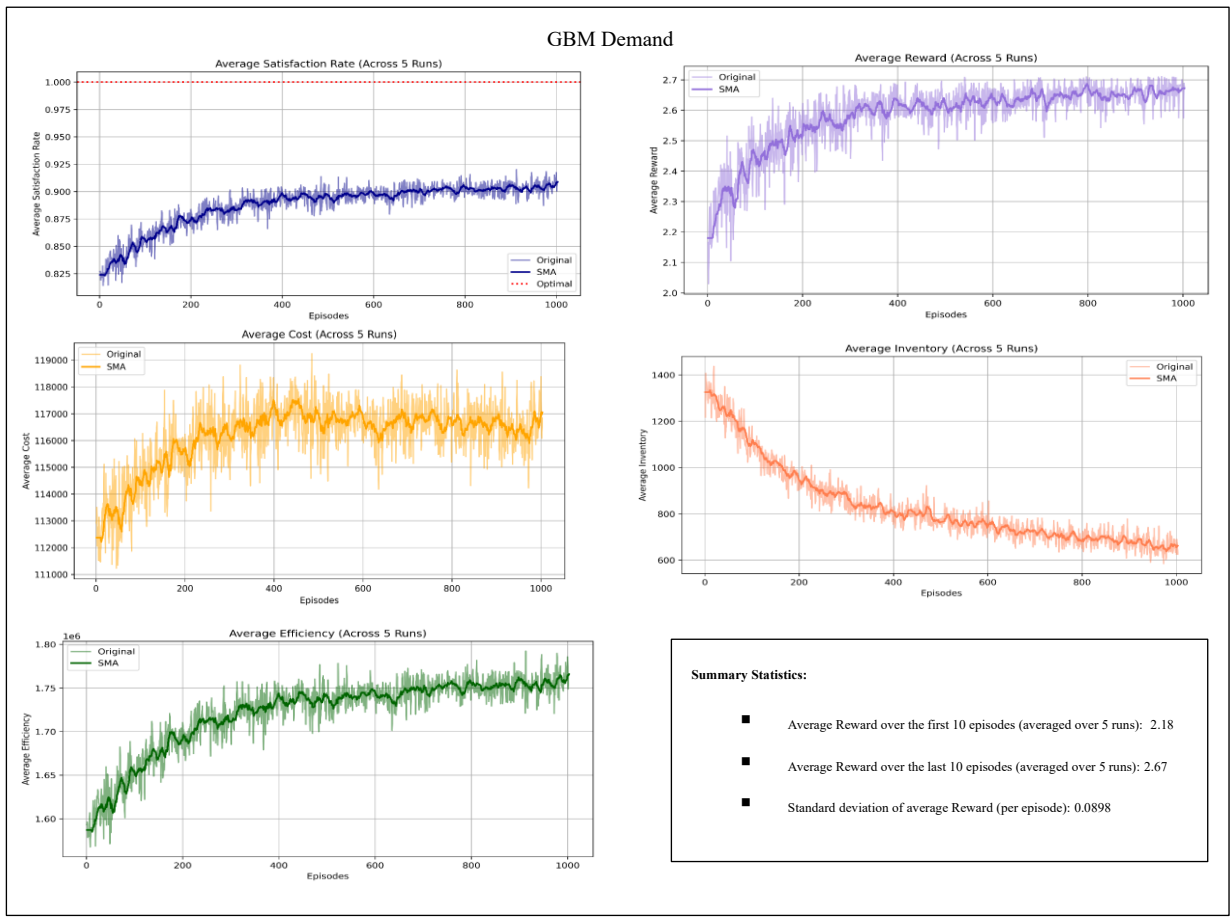

### B.4 Poisson demand (0-3000)

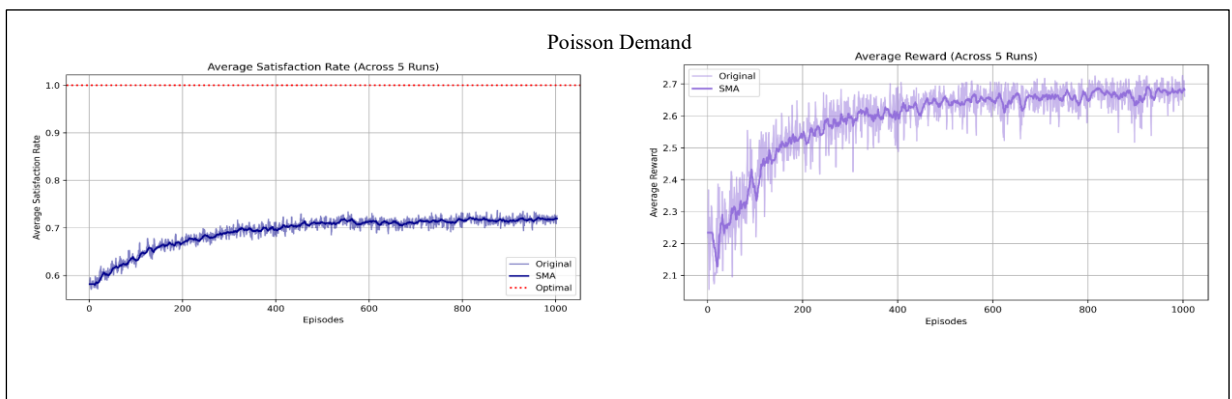

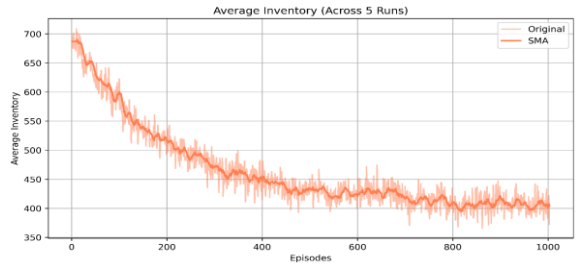

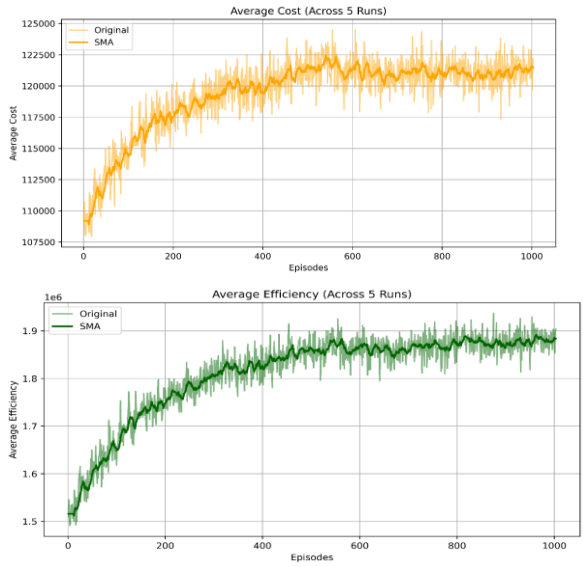

## B.5 GBM demand (0-1000)

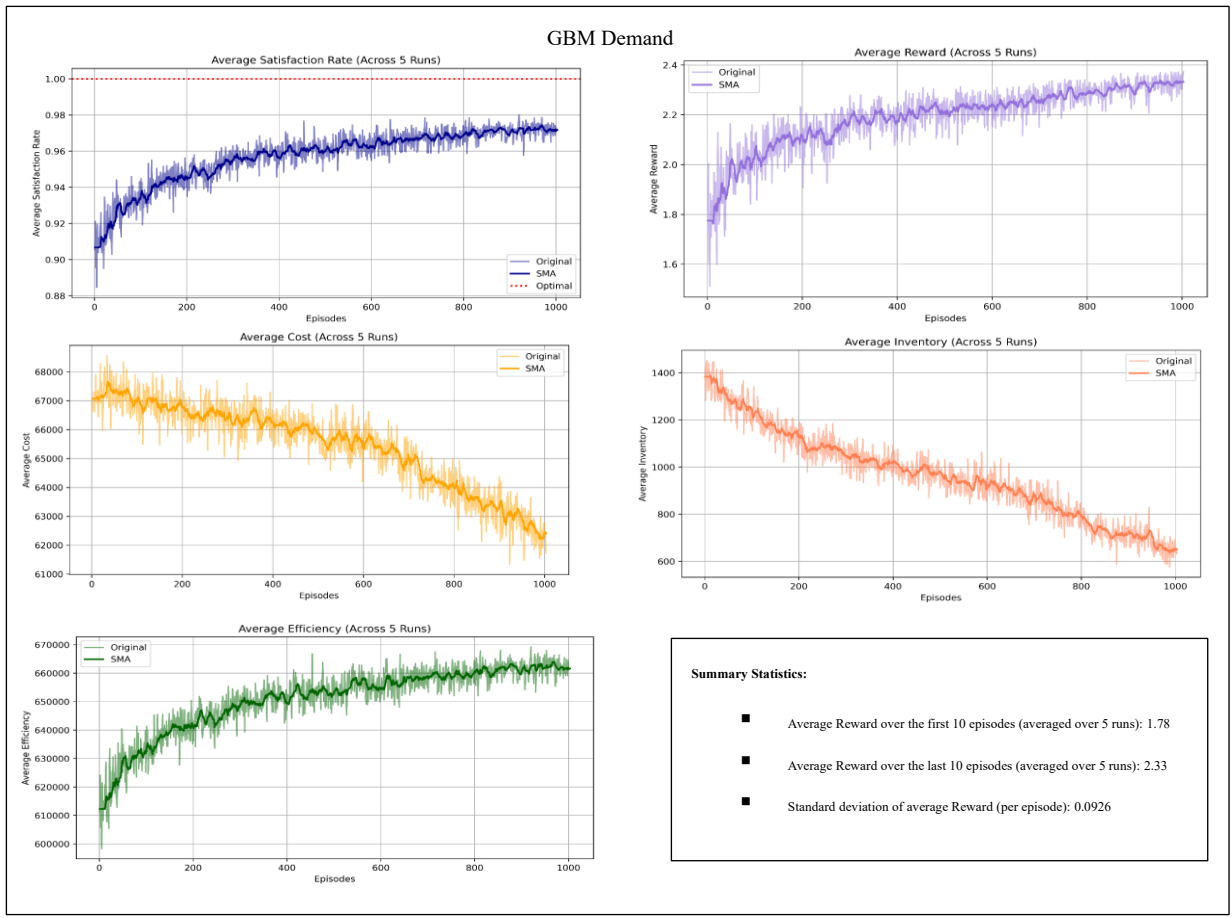

## B.6 Poisson demand (0-1000)

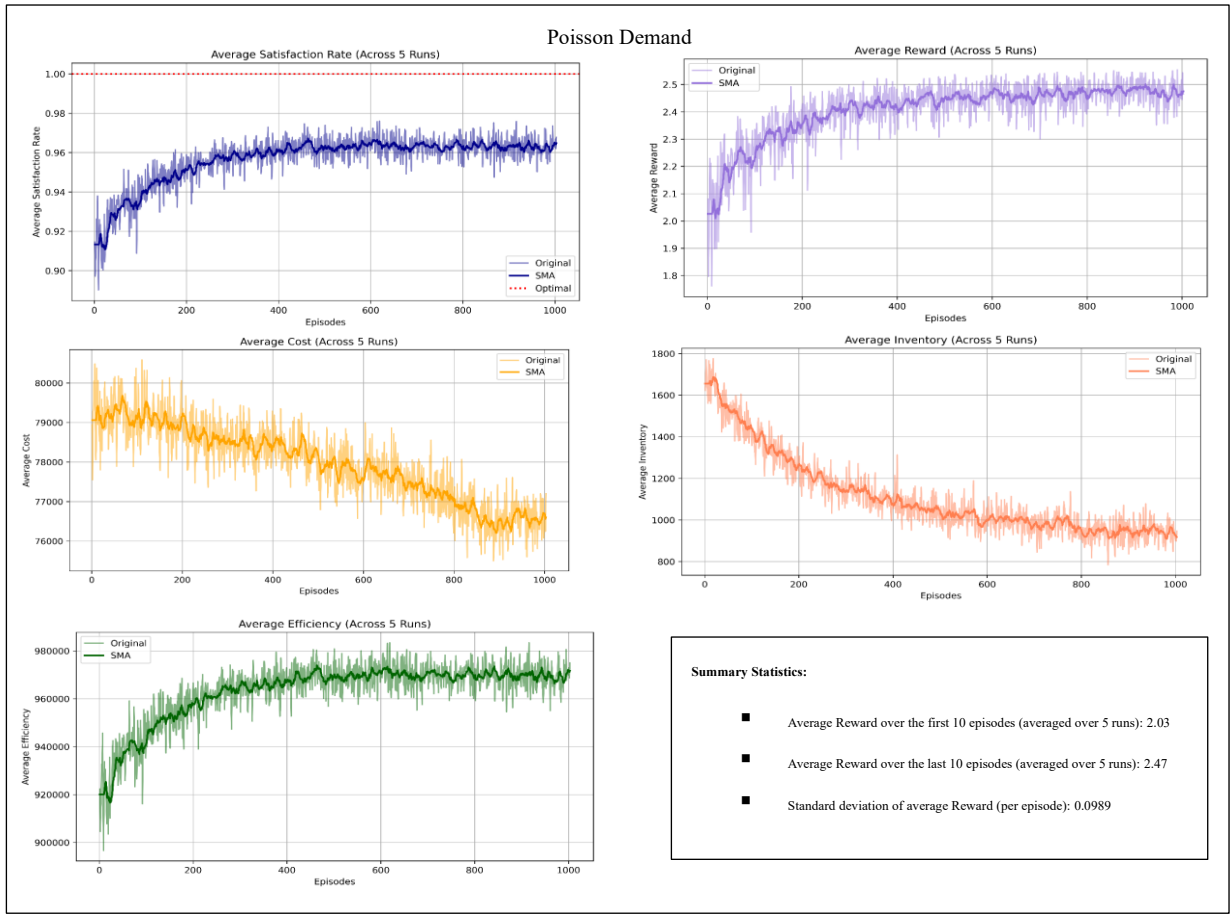

## B.7 Merton demand (0-1000)

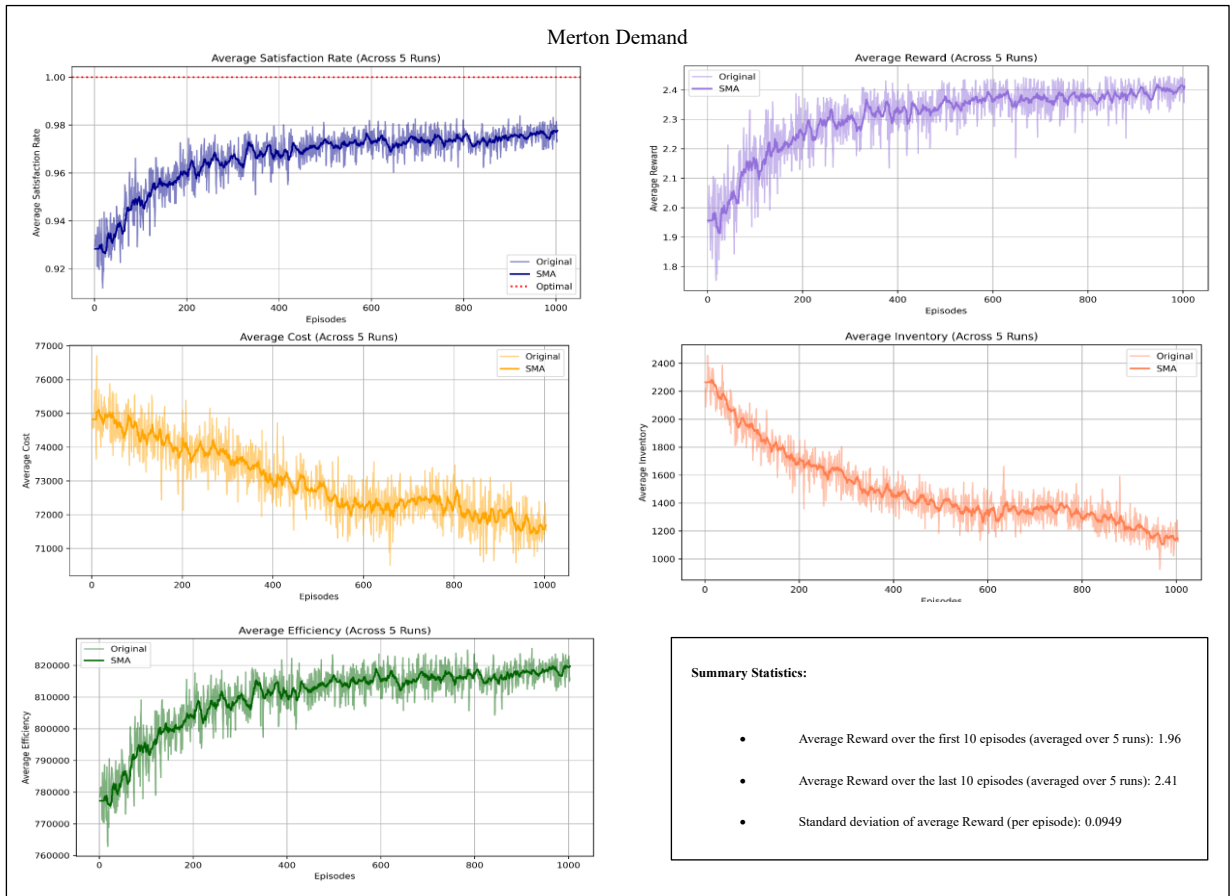

## B.8 Sensitivity test: Average Reward

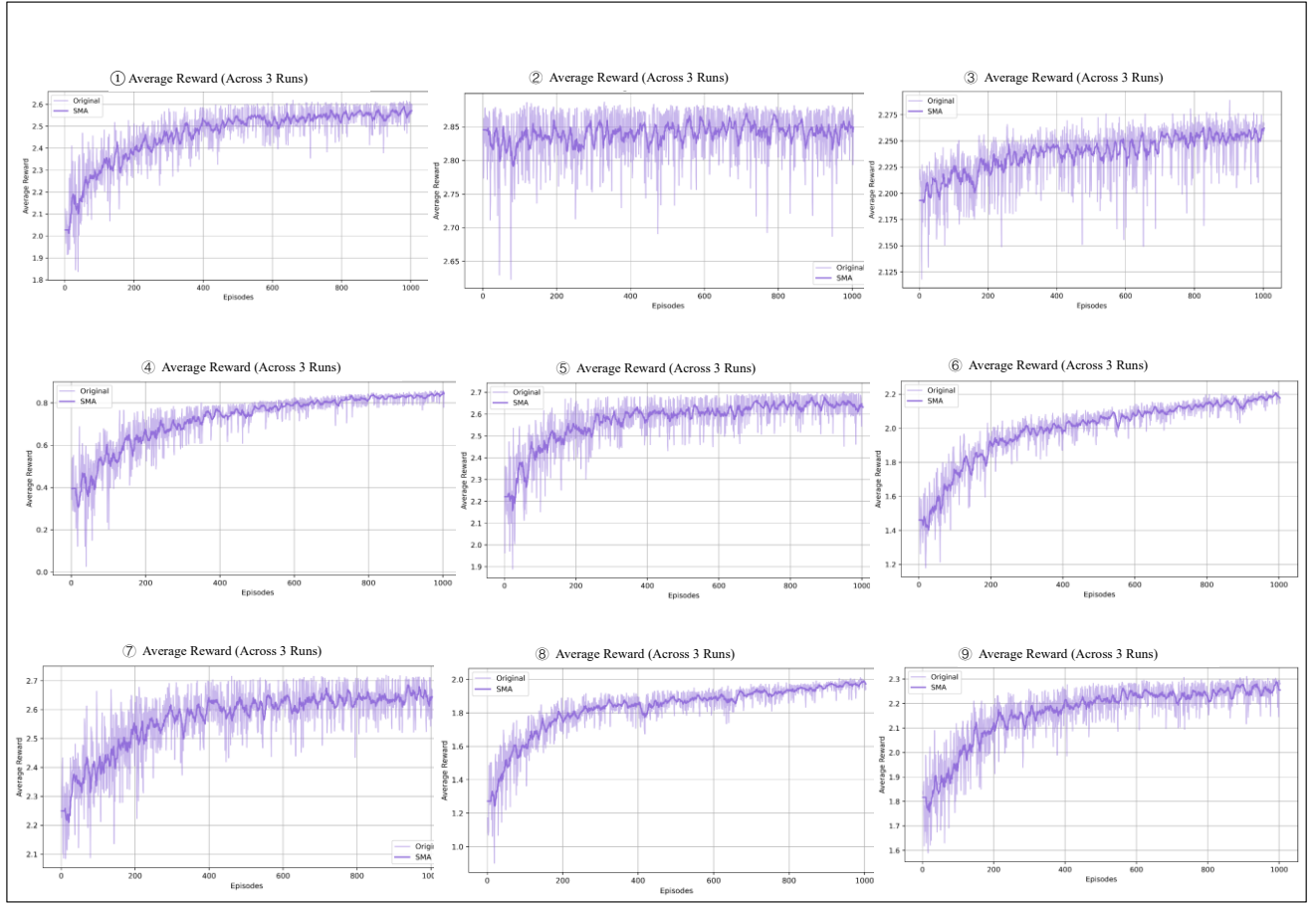

## B.9 Sensitivity test: Average Efficiency

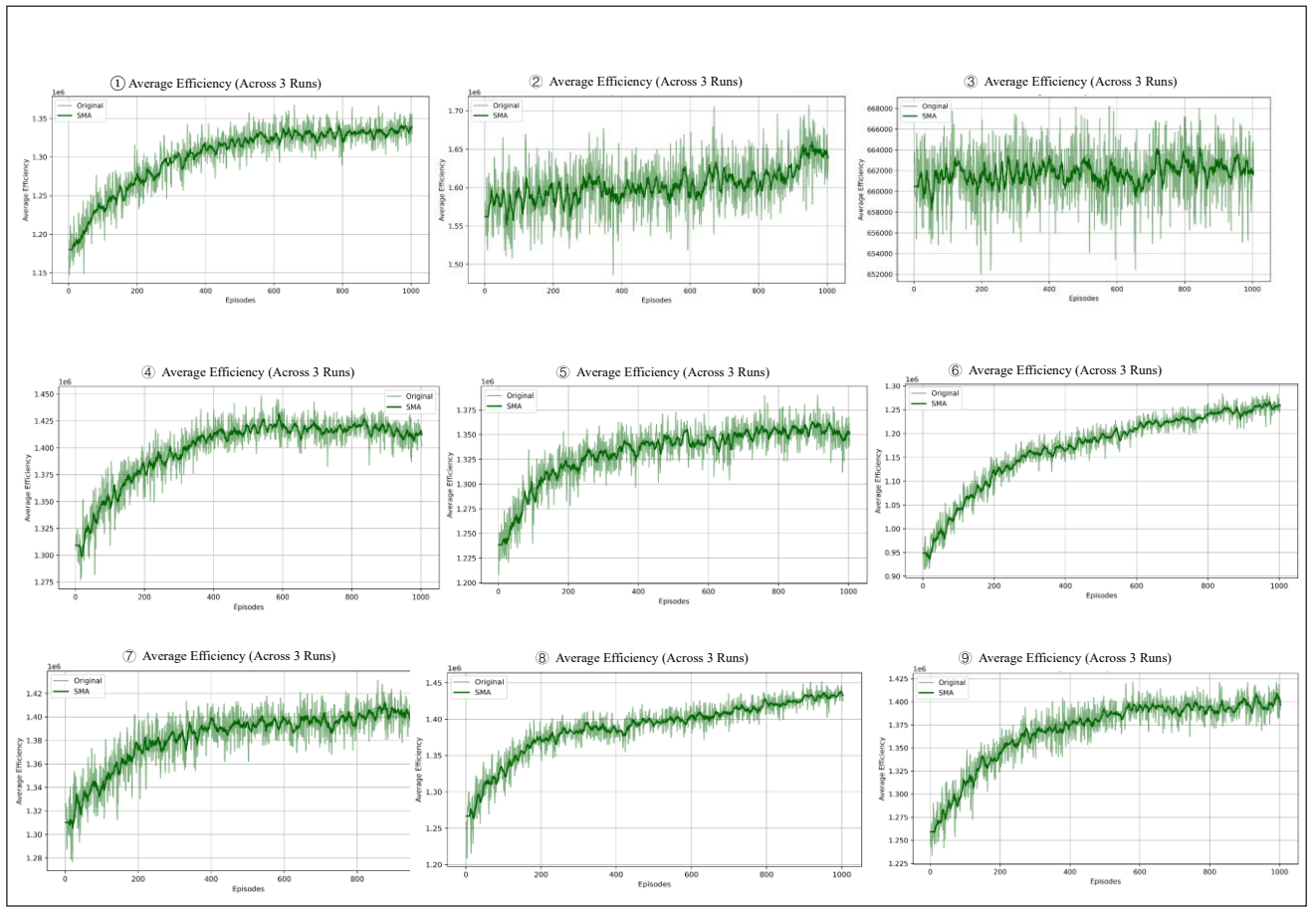

## B.10 Sensitivity test: Average Cost

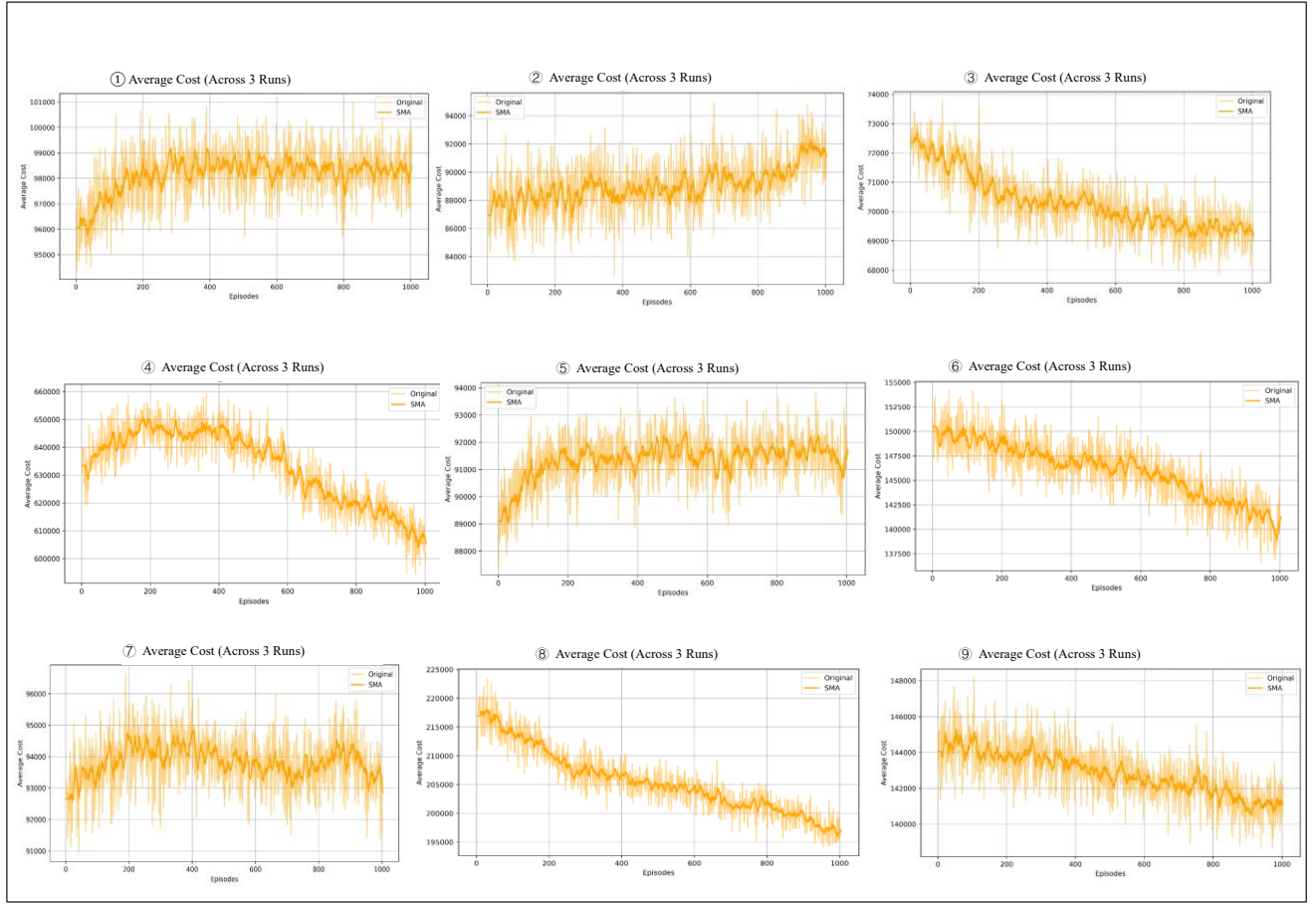

## B.11 Sensitivity test: Average Inventory

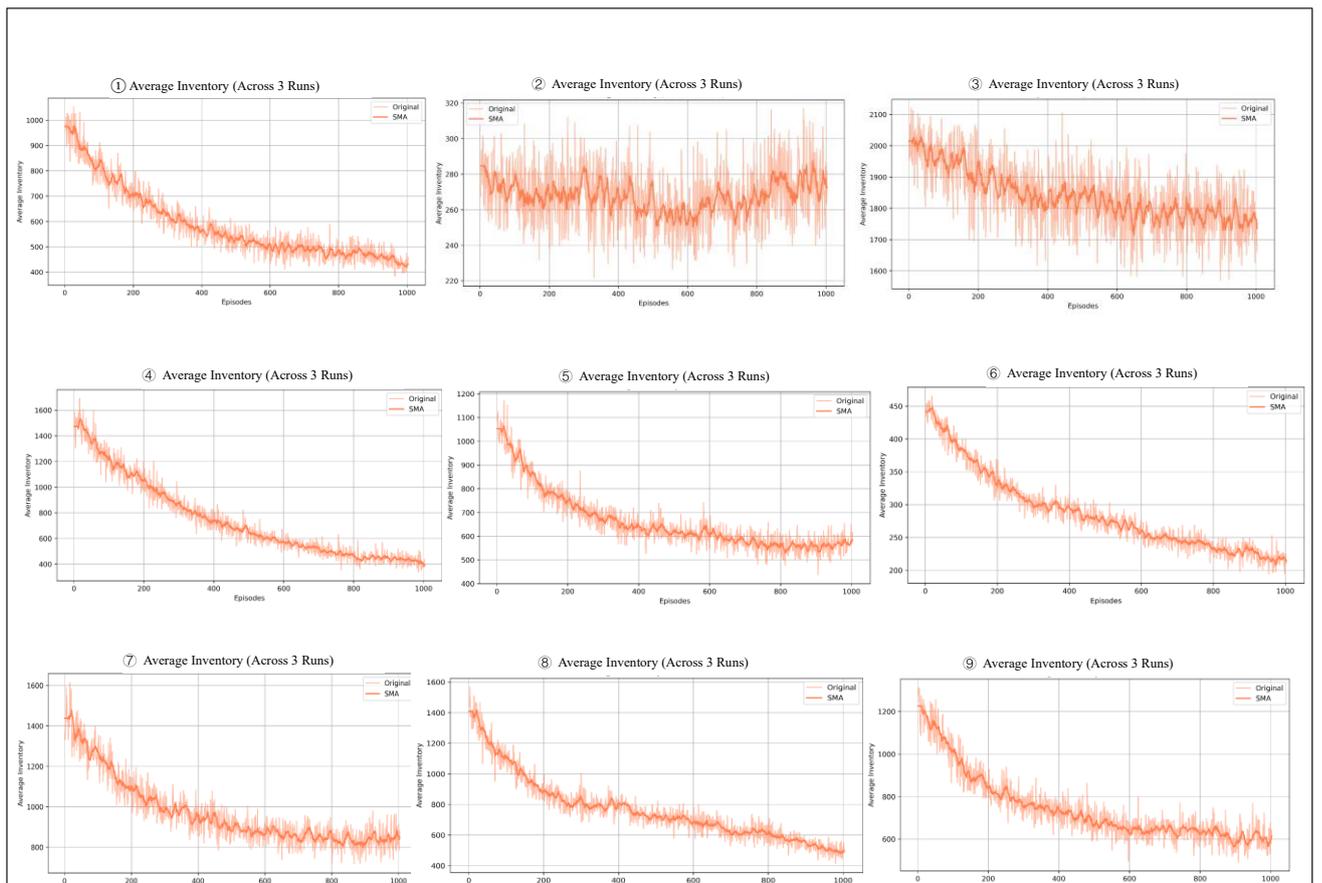

## B.12 The rest tests of impacts of $h$ and $V$ on average reward

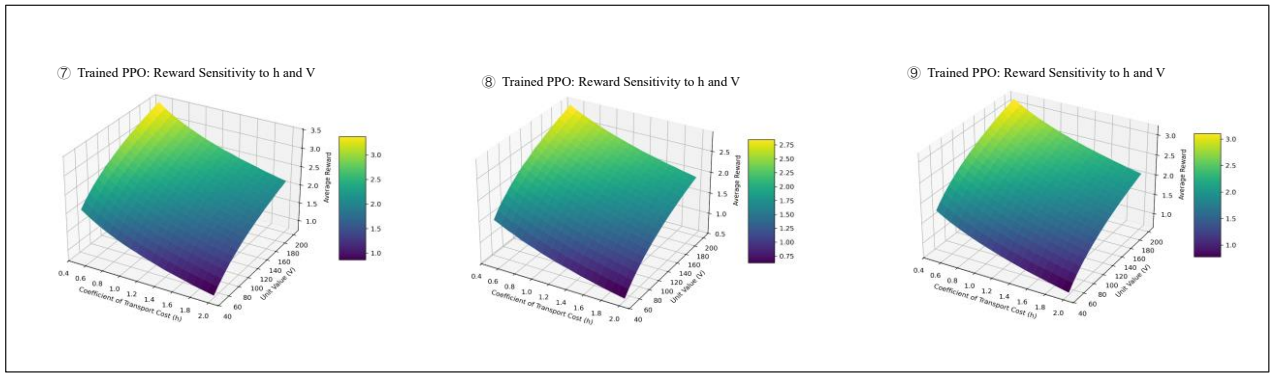

## B.13 Adjusted mismatching penalty with the consideration of pre-timestep inventory

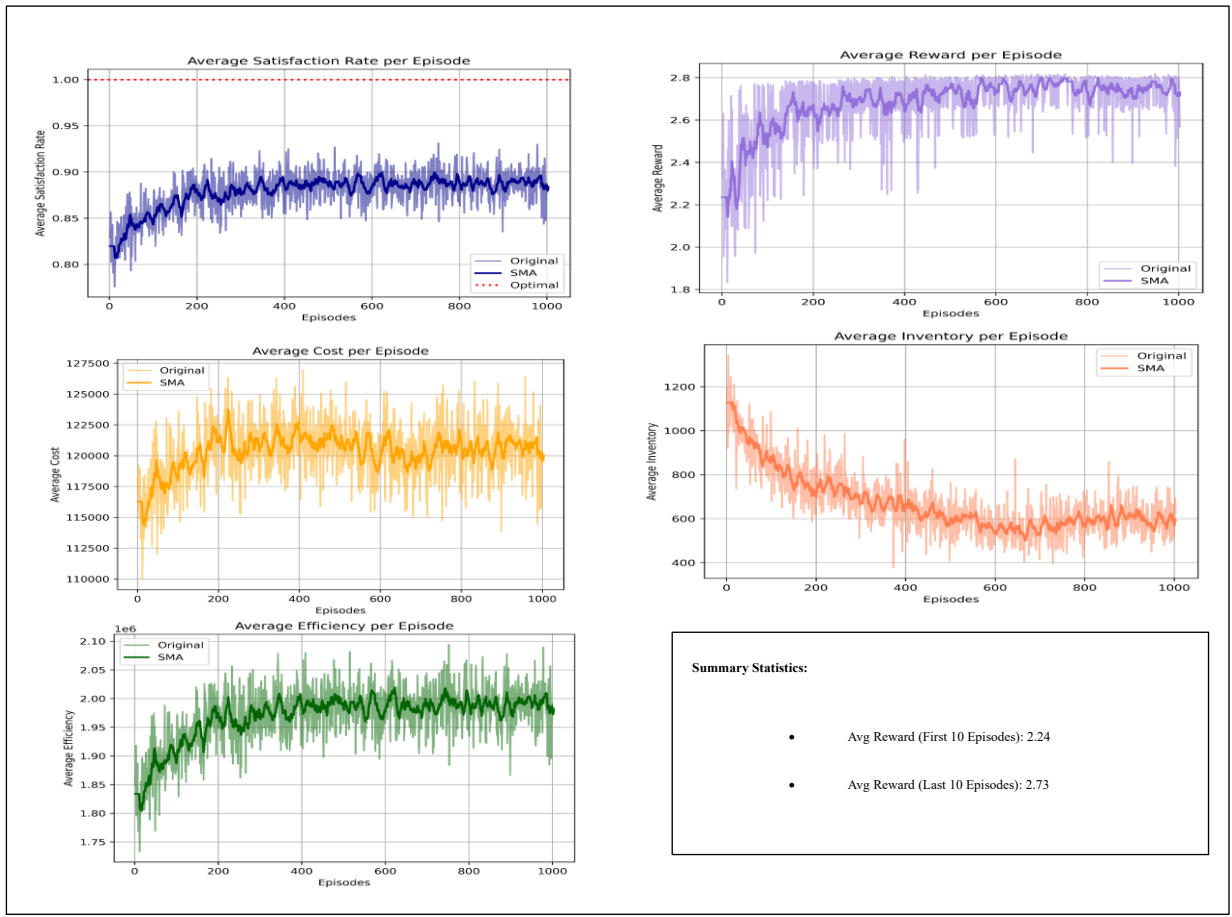

## B.14 Merton demand for four method comparison

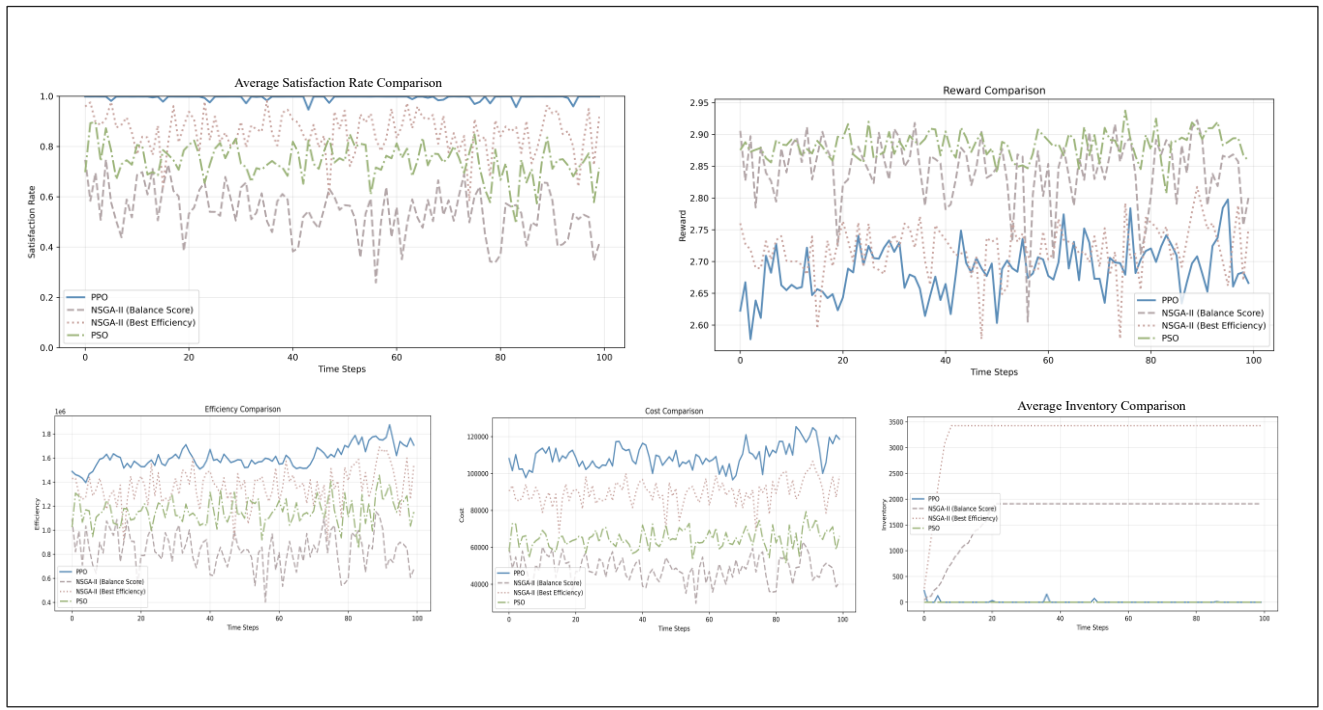

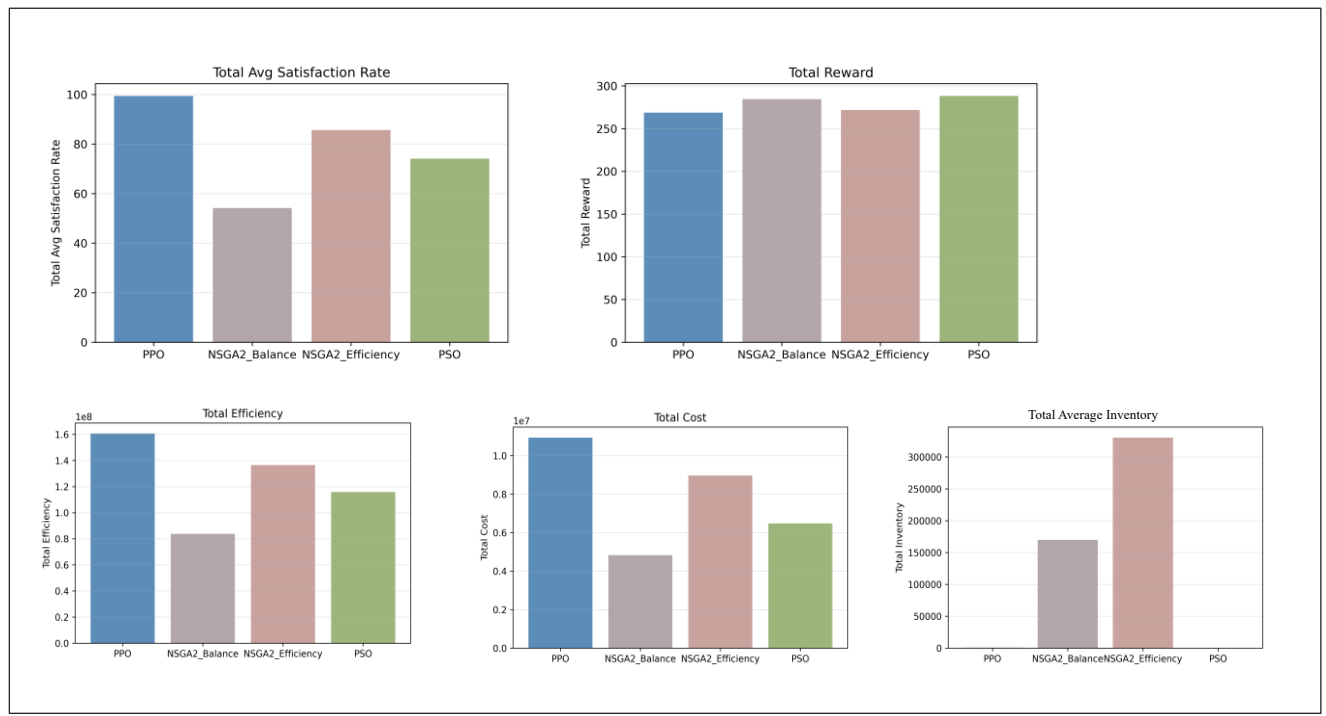

## B.15 Poisson demand for the four method comparison

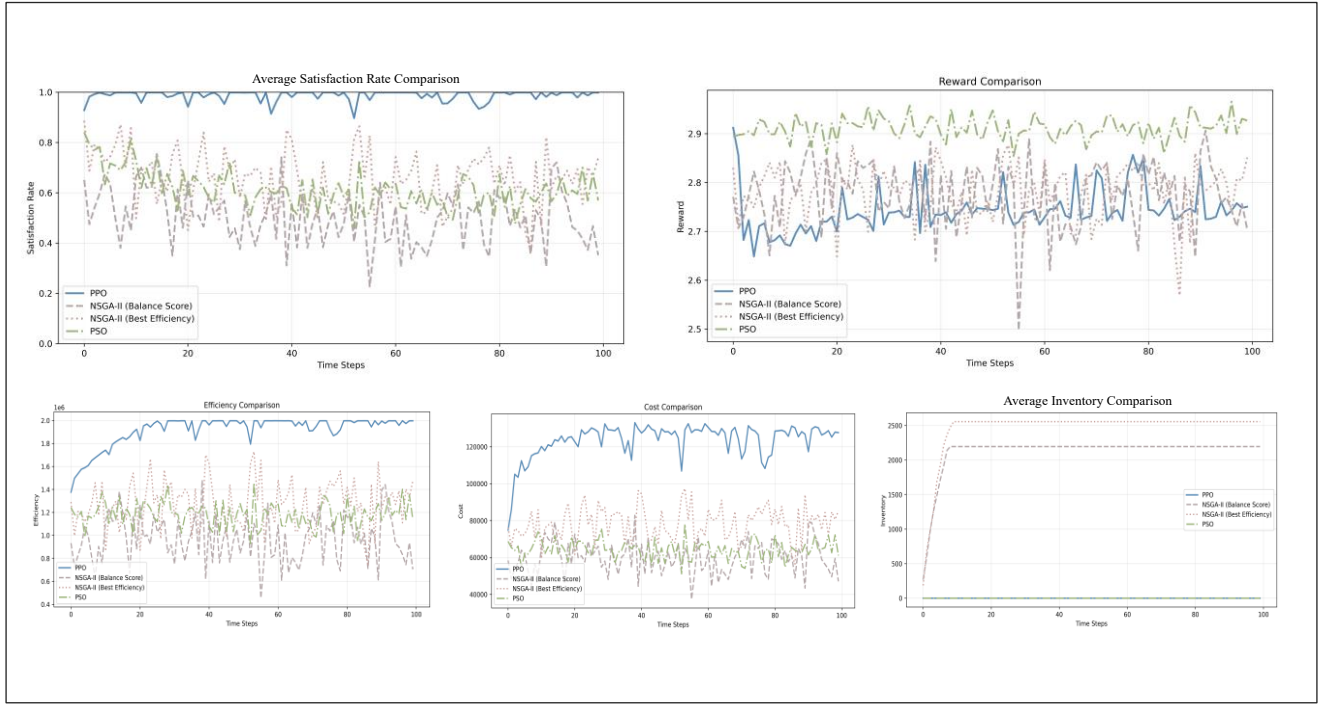

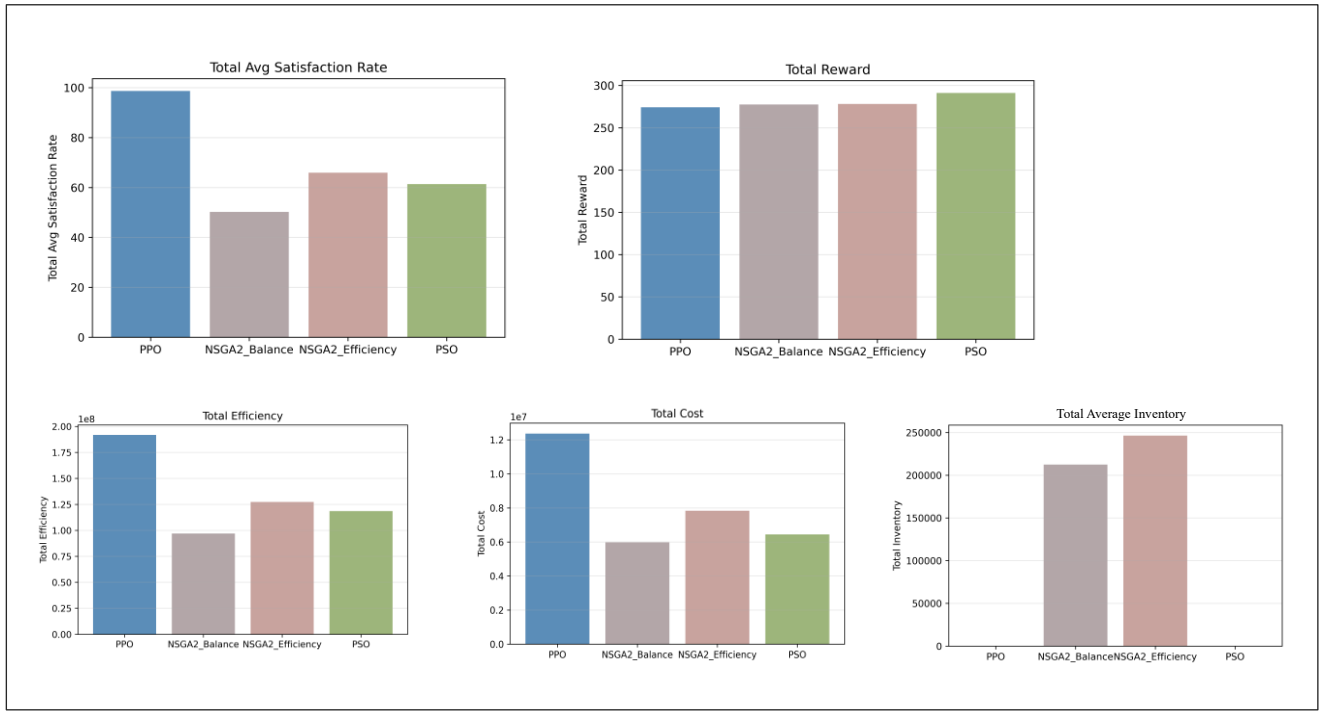

B. 17

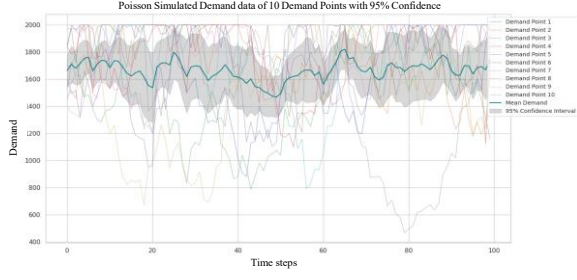

## C Pseudocode of PPO Key Implement

Initialize policy $\pi_0$ with parameters $\theta_0$
Set exploration rate $e_0$
Initialize reward log R = []
for k = 0 to N_iter do
    Initialize experience buffer B = ∅

    for m = 0 to N_episode do
        Reset environment: $I_w^t = 0$
        Observe initial state $s_0 = [d_p^0, I_w^0, A_{wp}^{-1}(None), 0]$

        for $t = 0$ to $T - 1$ do
            Select action $a_t \sim \pi_\theta(s_t)$
            Execute action $a_t$:
                - $x_c^t, x_w^t$
                - Simulate the distributions of collection centers as formula (16)
                - Simulate distributions of warehouses as formula (17)
                - Scale $A_{cw}^t$ to avoid over-supply as formular (20)

        Update warehouse inventory as formulas (18), (19)

        Compute unmet demand: $u_p^t = \max(d_p^t - \sum A_{wp}, 0)$

        Compute satisfaction rate: $r_p^t = \frac{A_{wp}^t}{d_p^t}$, $\bar{r}^t = \frac{1}{|P|}\sum_{p \in P} r_p^t$

        Compute efficiency as the formula (5)

        Compute cost as the formula (6)

        Compute penalties as the formulas (7), (8), (9).

Compute reward as the formula (4)

            Store $(s_t, a_t, R_t, s_{t+1})$ into buffer B
            Update state $s_t \leftarrow s_{t+1}$
        end for
    end for

    Update policy $\pi_\theta$ via PPO using buffer B
    Append mean reward to R
end for